\documentclass[nohyperref]{article}

\usepackage{microtype}
\usepackage{graphicx}
\usepackage{subfigure}
\usepackage{booktabs} 

\usepackage{hyperref}

\usepackage[accepted]{icml2021}

\usepackage{amsmath}
\usepackage{amssymb}
\usepackage{mathtools}
\usepackage{amsthm}

\usepackage{booktabs} 
\usepackage{bm}  
\usepackage{amsfonts}
\usepackage{enumitem}
\usepackage{epstopdf,algorithm,algorithmic}
\usepackage{color}
\usepackage{graphicx}           
\usepackage{subfigure} 

\usepackage[algo2e,linesnumbered,vlined,ruled]{algorithm2e}

\usepackage[capitalize,noabbrev]{cleveref}

\usepackage{hyperref}
\input{mysymbol.sty}

\theoremstyle{plain}
\newtheorem{theorem}{Theorem}[section]

\newtheorem{lemma}[theorem]{Lemma}

\theoremstyle{definition}

\newtheorem{assumption}[theorem]{Assumption}
\theoremstyle{remark}

\newtheorem{example}{\bf Example}

\def \EE {{\mathbb{E}}}

\def \e {{\varepsilon}}

\usepackage[textsize=tiny]{todonotes}

\icmltitlerunning{Policy Mirror Ascent in Continuous Action Spaces}

\begin{document}

\twocolumn[
\icmltitle{On the Hidden Biases of Policy Mirror Ascent in Continuous Action Spaces}



\icmlsetsymbol{equal}{*}

\begin{icmlauthorlist}
\icmlauthor{Amrit Singh Bedi}{umd}
\icmlauthor{Souradip Chakraborty}{umd}
\icmlauthor{Anjaly Parayil}{mct}
\icmlauthor{Brian Sadler}{arl}
\icmlauthor{Pratap Tokekar}{umd}
\icmlauthor{Alec Koppel}{amz}
\end{icmlauthorlist}

\icmlaffiliation{umd}{Department of Computer Science, University of Maryland, College Park, USA. }
\icmlaffiliation{mct}{Microsoft Research, India (work completed while at U.S. Army Research Laboratory, Adelphi, MD, USA).}
\icmlaffiliation{arl}{U.S. Army Research Laboratory, Adelphi, MD, USA.}
\icmlaffiliation{amz}{Supply Chain Optimization Technologies, Amazon, Seattle, USA.}

\icmlcorrespondingauthor{}{}

\icmlkeywords{Machine Learning, ICML}

\vskip 0.3in
]



\printAffiliationsAndNotice{\icmlEqualContribution} 

\begin{abstract}
We focus on parameterized policy search for reinforcement learning over continuous action spaces. Typically, one assumes the score function associated with a policy is bounded, which {fails to hold even for Gaussian policies. } To properly address this issue, one must introduce an exploration tolerance parameter to quantify the region in which it is bounded. Doing so incurs a persistent bias that appears in the attenuation rate of the expected policy gradient norm, which is inversely proportional to the radius of the action space. To mitigate this hidden bias, heavy-tailed policy parameterizations may be used, which exhibit a bounded score function, but doing so can cause instability in algorithmic updates. To address these issues, in this work, we study the convergence of policy gradient algorithms under heavy-tailed parameterizations, which we propose to stabilize with a combination of mirror ascent-type updates and gradient tracking.  Our main theoretical contribution is the establishment that this scheme converges with constant step and batch sizes, whereas prior works require these parameters to respectively shrink to null or grow to infinity. Experimentally, this scheme under a heavy-tailed policy parameterization yields improved reward accumulation across a variety of settings as compared with standard benchmarks.
\end{abstract}

\section{Introduction}

In reinforcement learning (RL), an autonomous agent sequentially interacts with its environment and observes rewards incrementally across time \citep{sutton2017reinforcement}. This framework has been successfully applied in continuous control \citep{schulman2015high,lillicrap2016continuous}, web services \citep{zou2019reinforcement}, personalized medicine \citep{kosorok2015adaptive}, among other contexts. Mathematically, RL may be defined by a Markov Decision Process (MDP) \citep{puterman2014markov}, where an agent seeks to select actions to maximize the long-term accumulation of rewards, known as the value. The key distinguishing point of RL with classical optimal control is its ability to discern policies without a system dynamics model. 

Algorithms for RL may be categorized as those which operate by approximately solving Bellman's equations \citep{bellman57a,watkins1992q} and policy gradient (PG) methods \citep{williams1992simple}. While the former may be lower variance and converge faster \citep{even2003learning,devraj2017zap}, typically they require representing a $Q$-function for every state-action pair, which is intractable for large spaces, the focus of this work. For this reason, we focus on PG methods. 

Policy search hinges upon the Policy Gradient (PG) Theorem \citep{sutton2000policy}, which expresses the gradient of the value function with respect to policy parameters as the expected value of the product of the score function of the policy and its associated $Q$ function. Policy search has been classically studied from the perspective of dynamical systems \citep{kushner2003stochastic,borkar2008stochastic}, leading to an understanding of its asymptotic performance \citep{konda1999actor,konda2000actor,bhatnagar2009natural}).
More recently, its non-asymptotic performance has come to the fore. {Recent results \citep{bhandari2019global,zhang2020sample,agarwal2020optimality} have established the global convergence of PG methods but require either softmax or direct policy parametrization. One may further refine the convergence constants via proximal regularization  \citep{schulman2017proximal,tomar2020mirror,lan2021policy,khodadadian2021linear}. These results hold for finite state and action spaces. However, for continuous spaces  the focus of this work) or general parameterizations, these results do not apply. Instead, stochastic gradient iteration for general non-convex objectives defines its performance in terms of convergence to stationarity  \citep{bhatt2019policy,zhang2020global}, i.e., $\mathcal{O}(1/\sqrt{K})$ (or $\mathcal{O}(\epsilon^{-2})$) rate of decrease of the policy gradient norm\footnote{($\epsilon$-optimal solutions) In this work, we call $\bbtheta \in\mathbb{R}^d$ as $\epsilon-$optimal if and only if $\mathbb{E}[\|\nabla F(\bbtheta)\|]\leq \epsilon$. Sometimes, $\mathbb{E}[\|\nabla F(\bbtheta)\|^2]\leq \epsilon$ is used in literature which means that our $\mathcal{O}(\epsilon^{-4})$ would be equivalent to $\mathcal{O}(\epsilon^{-2})$.}}.

\begin{table}[]
	\centering
\resizebox{\columnwidth}{!}{	\begin{tabular}{|c|c|c|c|}
		\hline
		\multicolumn{1}{|c|}{Algorithms} & \multicolumn{1}{c|}{Hidden bias} & Bregman term & SC 
		\\
		 \hline
		\multicolumn{1}{|c|}{SVRPO \citep{xu2017stochastic}}      & \multicolumn{1}{c|}{Yes}            &   No         &   N/A 
		\\
		 \hline
	SVRPG \citep{papini2018stochastic}	&      Yes                          &    No       &   $\mathcal{O}(\epsilon^{-4})$
		 \\
		  \hline
		\multicolumn{1}{|c|}{STORM-PG \citep{yuan2020stochastic}}      &  \multicolumn{1}{c|}{Yes}            & No        &    $\mathcal{O}(\epsilon^{-4})$ 
		\\
		 \hline
RPG \citep{zhang2020global}		&    Yes                              & No         &    $\mathcal{O}(\epsilon^{-4})$
\\
 \hline
SRMA (This work)		&       \textbf{No}                           &   \textbf{Yes}         &   $\mathcal{O}(\epsilon^{-4})$ 
\\
 \hline
	\end{tabular}}
\caption{{This table summarizes the existing sample complexity results for the policy gradient algorithm in literature for continuous state-action spaces. We note that all the results present in in the table are modified  according to criteria $\mathbb{E}[\|\nabla F(\bbtheta)\|]\leq \epsilon$ and constant batch size of $\mathcal{O}(1)$. True to our knowledge, this work provides the first optimal sample complexity results with finite batch size for the policy gradient algorithm with Bregman divergence as the regularization term.}   }
\label{literature}
\end{table}

Noticeably, these results require the score function to be uniformly bounded, which fails to encapsulate even standard Gaussian parameterizations with finite variance. This issue is identified in \citep{bedi2021sample}, where in continuous space with an unbounded score function, instead the resultant rate is established as $\mathcal{O}(1/\sqrt{K}) + \mathcal{O}(\lambda)$. Here scalar $\lambda$ is the exploration tolerance which quantifies the radius of action space. So, for Gaussian distribution and related parameterized families, a tradeoff between exploring the state space via extreme actions and converging to stationarity is present. Ideally, we would like $\lambda$ to be very small, but doing so restricts the space over which action selection may take place (cf. \citep[Definition 4.1]{bedi2021sample}). To achieve better space coverage, one must allow greater range of actions, but this yields large $\lambda$, which degrades the tightness of the radius of convergence to stationarity. Therefore, we pose the following question in this work:
	
	``\textit{Which policy parameterizations achieve exact convergence to stationarity in continuous state and action space?}" 
	
	This work presents an affirmative answer to this question through the identification that score function boundedness is a property of heavy-tailed distributions such as the Cauchy \citep{hutchinson1981fractals,focardi2003fat,barabasi2003emergence}. This boundedness then mitigates the bias associated with the exploration tolerance parameter. Moreover, heavy-tailed policies in practice has been shown to yield improved state space coverage and escape spurious extrema in continuous space in \citep{chou2017improving,papini2020balancing,garg2021proximal}. For these reason, we focus on policy search under heavy-tailed policy parameterizations \footnote{As a side note, heavy-tailed policies induce heavy-tailed gradient noise, which has been associated with improved generalization in supervised learning recently \citep{NEURIPS2020_37693cfc,simsekli2020fractional}; however, we defer a rigorous exploration of this phenomenon in RL to future work.}.
{Substituting a sub-Gaussian distribution with a heavy-tailed one, however, may cause instability in the parameter estimates due to increased likelihood of extreme action selection \citep{nemirovski2009robust}. To ameliorate this issue, we prioritize proximal variants of PG \citep{schulman2017proximal,tomar2020mirror}, which employ Bregman divergence as a regularization. The idea of proximal algorithm was first introduced in \cite{rockafellar1976monotone}, and then generalized to Bregman divergences in \cite{nemirovski2009robust} for convex objectives. Later, mirror descent approach was extended to non-convex objectives in \cite{ghadimi2016mini} and shown to converge for increasing batch sizes. In the RL literature, the global convergence of policy mirror ascent has been established recently \citep{tomar2020mirror,khodadadian2021linear,lan2021policy}, but demands that one regularize the value function, and send the step-size or batch size to null or infinity \citep{ghadimi2016mini}, respectively, in order to ensure convergence. In this work, we address this impracticality by introducing a gradient tracking mechanism that is able to applicable to operate with finite mini-batch sizes, which we call Stochastic Recursive Mirror Ascent Algorithm (SRMA). SRMA is inspired by \citep{cutkosky2019momentum}, but differs in a key ways. In particular, we develop a gradient update that employs importance sampling to address distributional shift inherent to RL \citep{yuan2020stochastic}; { \bf here, however, it is used in a novel context, which is to reduce persistent errors in the Bregman gradient, rather than for the purpose of variance reduction}. This point of departure is underscored by the fact that both \citep{cutkosky2019momentum,yuan2020stochastic} do not use proximal updates, and do not characterize how the convergence rate depend on the policy parameterization. Doing so then overlooks the hidden bias of PG in continuous space. A summary of related results in the literature is given in Table \ref{literature}. Therefore, our main contributions are as follows:}



 \begin{itemize}
 	\item We use an often overlooked fundamental theoretical gap regarding the boundedness of the norm of score function in policy gradient method in continuous spaces to rigorously motivate the use of heavy-tailed parameterizations, such as the Cauchy policy, which are not afflicted by hidden bias. 
	
	\item We develop a novel variant of policy mirror ascent that employs gradient tracking in its inner loop, which we call Stochastic Recursive Mirror Ascent (SRMA) (Algorithm \ref{Algorithm2}). This scheme additionally employs importance sampling to correct for distributional shift that manifests when one tries to develop momentum techniques in RL.
 	
 	\item Our main theoretical result is the establishment of convergence to stationarity (Theorem \ref{main_khorem}) of SRMA, which contrasts with prior results on stochastic mirror ascent that demand increasing batch sizes or decreasing step-sizes in order to obtain convergence, and closes a conspicuous gap in the literature for proximal methods applied to non-convex expected value objectives \emph{without regularization under constant step-size} -- see \citep{lan2021policy} for a study of the finite-state case. The convergence proof is done with a general set of assumptions which are explicitly proved to holds for the RL setting considered in this paper. 
 	
 	 
 	\item We provide extensive simulations to show that policies learned with heavy-tailed distributions perform favorably in RL problems where biases associated with insufficiently covering the state spaces may be present (Sec. \ref{sec:simulations}).
 \end{itemize}



\section{Markov Decision Problems}\label{sec:prob}
In reinforcement learning (RL), an autonomous agent traversing through a state space $\mathcal{S}$ at state $s$, selects action $a\in\mathcal{A}$ and transitions to another state $s'$ according to a Markov transition density $\mathbb{P}(s'|s,a)$. Upon reaching state $s'$, the environment reveals an instantaneous reward $r(s,a)$ which informs the merit of a given decision $a$ starting from state $s$. Mathematically, this framework for interactive decision-making may be defined as a Markov Decision Process (MDP), whose components are  $(\mathcal{S}, \, \mathcal{A}, \, \mathbb{P},\,r,\, \gamma )$. The state  $\mathcal{S}$ and action space $\mathcal{A}$ may either be finite or compact real vector space such that $\mathcal{S} \subseteq \mathbb{R}^{q}$ and $\mathcal{A} \subseteq  \mathbb{R}^{p}$. Moreover, $\gamma$ is a discount factor that determines how much future rewards are worth relative to the next step.  As is well known in MDPs \citep{bertsekas2004stochastic,puterman2014markov}, it suffices to hypothesize the decision-maker selects actions $a_t\sim \pi(\cdot |s)$ over a time-invariant distribution $\pi(a|s):= \textrm{Pr}\{{a}_t=a| s_t=s \}$ called a policy, which denotes the  probability of action $a$ given the agent is in state $s$.  The goal in RL is to determine the policy that accumulates the most long-term reward on average, i.e., the value:
\begin{align}\label{eq:value_func}
V^{\pi}(s) = \mathbb{E} \bigg[ \sum_{t=0}^{\infty} \gamma^{t} r_t~|~s_0=s, a_t= \pi(s_t)  \bigg], 
\end{align}
where $s_0$ denotes the initial state along a trajectory $\{s_t,a_t,r_t\}_{u=0}^\infty$ with short-hand notation $r_t=r(s_t,a_t)$. Here, the expectation in \eqref{eq:value_func} is with respect to the product measure of randomized policy $a_t \sim \pi(\cdot |s_t)$ and state transition dynamics $s_{t+1} \sim \mathbb{P}(.|s_t,a_t)$. For further reference, we  define the action-value, i.e., $Q$-function as the value conditioned on an initially selected action as $Q^{\pi}(s,a)$$=$$\mathbb{E}\big[\sum_{t=0}^{\infty}\gamma^{t}r_{t}|s_0=s,$$ a_0=a,$$ a_t$$=$$\pi(s_t)\big]$.
Our focus is on policy search over parameterized families of policies, which hypothesizes that actions are selected according to a policy $ \pi_{\bm{\theta}}(\cdot |s_t)$ parameterized by vector $\bm{\theta} \in \mathbb{R}^d$. Then, we seek to estimate those parameters that maximize the cumulative return \citep{sutton2017reinforcement} given by
\begin{align}\label{eq:main_prob}
\max_{\bm{\theta}} J(\bm{\theta})
\end{align}
where, objective is given by $J(\bm{\theta}) : = V^{{\pi}_{\bm{\theta}}}(s_0)$. Observe that \eqref{eq:main_prob} is non-convex in $\bm{\theta}$, and therefore, finding the optimal policy is challenging even in the deterministic setting. However, in RL, the search procedure necessarily interacts with the transition dynamics $\mathbb{P}(s'|s,a)$ as well. Before detailing how one may implement first-order stochastic search to solve \eqref{eq:main_prob}, we introduce the widely used standard Gaussian policy parameterization, and clarify how its practice can lead to hidden bias.

%

\begin{example}[Gaussian Parametrization]\normalfont \label{eg:gaussian_fixed_var}
For continuous spaces, the Gaussian policy takes the form
%
%
$\pi_{\bm{\theta}}(a|s) = \mathcal{N} (a|\varphi(s)^\top \bm{\theta}, \sigma^2)$,
%
%
where the parameters $\bm{\theta}$ determine the mean (centering) of a Gaussian distribution at $\varphi(s)$, and $\sigma^2$ is a fixed-variance hyper-parameter. Here, $\varphi(s)$ denotes the state space feature map, i.e., $\varphi: \mathcal{S}\rightarrow \mathbb{R}^d$ with $d\ll q$.
\end{example}

\subsection{Unbounded Score Functions and Hidden Bias}\label{section:algorithm}
Policy gradient (PG) method is an algorithm for RL which operates by implementing approximate gradient ascent in parameter space $\bbtheta \in \mathbb{R}^d$ with respect to the value function \eqref{eq:value_func}. The key enabler of this method is the Policy Gradient Theorem \citep{sutton2017reinforcement}, which expresses search directions in parameter space as
\begin{align}
	\nabla J(\theta)
	&=\frac{1}{1-\gamma}\cdot\EE\big[\nabla\log\pi_{\theta}(a\given s)\cdot Q^{\pi_\theta}(s,a)\big], \label{eq:policy_grad}
\end{align}
where the expectation is over $(s,a)$$\sim $$\rho_{\theta}(\cdot,\cdot)$ and  $\rho_{\theta}(s,a)$$=$$\rho_{\pi_\theta}(s)\cdot \pi_{\theta}(a\given s)$ is a probability distribution that denotes the \emph{discounted state-action  occupancy measure}, which is the product of the discounted state occupancy measure $\rho_{\pi_\theta}(s)$$=$$(1-\gamma)\sum_{t=0}^\infty\gamma^t \mathbb{P}(s_k=s\given s_0,\pi_\theta)$ and  policy $\pi_{\theta}(a\given s)$. In \citep{sutton2000policy}, both $\rho_{\pi_\theta}(s)$ and $\rho_{\bm{\theta}}(s,a)$ are established as valid probability distributions. 
To compute policy search directions, we consider an unbiased estimator of policy gradient via randomized horizon $T_k\sim\text{Geom}(1-\gamma^{1/2})$ with trajectory $\xi_k(\bbtheta_k)=\{(s_0,a_0)\cdots(s_{T_{k}},a_{T_{k}})\}$ given by
\begin{align}\label{eq:policy_gradient_iteration}
	{\nabla}  J(\bm{\theta}_k,&\xi_k(\bbtheta_k))\\
	=& \sum_{t=0}^{T_k}\gamma^{t/2}r_t\cdot\bigg(\sum_{\tau=0}^{t}\nabla\log\pi_{\bm{\theta}_k}(a_{\tau}\given s_{\tau})\bigg),\nonumber
\end{align}
such that stochastic gradient is unbiased (proof is available in \citep[Lemma 1]{bedi2021sample})  and $\xi_k(\bbtheta_k)$ is the randomness in the stochastic gradient estimate at $k$. This estimator is a variant of the one proposed in \cite{baxter2001infinite} but with a randomized horizon. This is in contrast to the existing literature where a fixed horizon length $T_k=H$ for all $k$ is utilized (see \cite{papini2018stochastic,yuan2020stochastic,xu2017stochastic}). We remark here that a fixed horizon $H$ actually leads to a bias-variance tradeoff in the gradient estimation rather than providing an unbiased estimate as discussed in \cite{baxter2001infinite}. Such tradeoff is not present for a randomized horizon based-estimator mentioned in \eqref{eq:policy_gradient_iteration} as it is unbiased, which is our motivation for using it in this work.  Here, $\xi_k(\bbtheta_k)$ is the trajectory collected by using $\bbtheta_k$ as the policy parameter and hence it is function of $\bbtheta_k$.  Note the summation in \eqref{eq:policy_gradient_iteration} over two index: index $t$ denotes rollout trajectory information and $\tau$ is collecting the score function from starting to current index $t$. Using this scheme, stochastic policy gradient method iterates as
\begin{align}\label{eq:policy_gradient_iteration22}
	\bm{\theta}_{k+1} =& \bm{\theta}_k + \eta {\nabla} J(\bm{\theta}_k, \xi_k(\bbtheta_k)), \; 
\end{align}
where  $\eta>0$ denotes the step size.  
%
%
For general parametrized policy $\pi_{\bm{\theta}}$, by employing the iteration \eqref{eq:policy_gradient_iteration22}, one may obtain convergence to stationary points of \eqref{eq:main_prob} (see \citep{zhang2020global,zhang2020sample}).
To do so, however, to date ({see list in Table \ref{literature}}), most results require the score function to be deterministically bounded over the entire state space and action space. Unfortunately, this assumption is violated for the basic Gaussian policy (Example \ref{eg:gaussian_fixed_var}). To emphasize this point, note that score function, for scalar action space for simplicity, associated with the Gaussian policy (cf. Example \ref{eg:gaussian_fixed_var}) is 
\begin{align}\label{first_gradient}
& \nabla \log \pi_{\bm{\theta}} (s,a)= \frac{(a-\varphi(s)^\top\bm{\theta})\varphi(s)}{\sigma^2},
\end{align} 
where $\varphi(s)$ is the state features such that $\|\varphi(s)\|\leq D$ for all $s$.  \eqref{first_gradient} makes clear that the score function norm upper bound $|\nabla \log \pi_{\bm{\theta}} (s,a)|$$\leq$$ \mathcal{O}\left(D |a| \cdot +D^2\|\bbtheta\|\right)$ is linear with respect to $|a|$ and $\|\bbtheta\|$, which is unbounded unless the action space is compact.  However, a Gaussian distribution is only valid over \emph{infinite range}, leading to the following technical subtlety to address this issue. We note that identical logic applies in higher dimensions.

{\bf Exploration Tolerance and Hidden Bias} \citep[Def. 4.1]{bedi2021sample}:  Define $\mathcal{A}(\lambda)$ as the set of subsets of action space such that the score function has finite integral less than a scalar $\lambda>0$ with respect to the policy
	\begin{align}\label{define_Set}
\!\!\!\!\!		\mathcal{A}(\lambda):=\big\{\mathcal{C}\subseteq\mathcal{A}\!:\!\!\!\int_{\mathcal{\mathcal{A}\backslash\mathcal{C}}}\hspace{-5mm}\|\nabla\log\pi_{{\bbtheta}}(a | s)\| \pi_{\bbtheta}(a|s) da \leq \lambda\big\},
	\end{align}
for all $s$ and $\bbtheta$. $\lambda$ is the exploration tolerance parameter induced by a policy in an MDP whose score function is unbounded. In most existing analyses of PG algorithms, this quantity is ignored, leading to to a misconception that PG methods  converge exactly in expectation to a stationary point \citep{bhatt2019policy,zhang2020global}; however, when one properly accounts for $\lambda$, an $\mathcal{O}(\lambda)$ term appears as a persistent bias in the radius of convergence, i.e., PG in continuous space with a Gaussian policy yields an $\mathcal{O}(1/\sqrt{K} +\lambda)$-suboptimal policy after $K$ trajectories -- see \citep{bedi2021sample}[Theorem 4.2]. For Gaussian policy, to make $\lambda$ small, one must choose a large $\mathcal{C}$ (set of possible actions), which is usually restricted by the practical physical limitations such as the bounded acceleration of a vehicle. By contrast, heavy-tailed distributions \citep{hutchinson1981fractals,focardi2003fat,barabasi2003emergence}, for instance, the Cauchy, do not suffer this drawback. That is, their associated score functions are bounded. 	
$r(s_t,a_t;\bbtheta_{:t})= $
	\begin{example}[Cauchy Parametrization]\normalfont \label{eg:alpha_stable}
		%
		Symmetric $\alpha$ stable, $\mathcal{S}\alpha\mathcal{S}$ distributions are a generalization of a centered Gaussian distribution with $\alpha \in  (0, 2]$  as the tail index  which determines the heaviness of the distribution's tail  \citep{nguyen2019first}. Denote random variable $\textbf{X} \sim \mathcal{S}\alpha\mathcal{S} (\sigma)$ with associated characteristic function $\mathbb{E}\left[ e^{i\omega \textbf{X}}\right]  = e^{-| \sigma|\omega^{\alpha}}$ and scale parameter $\sigma  \in (0, \infty) $. Note that for
		$\alpha=1$ we have a Cauchy distribution given by
		\begin{align}\label{eq:Cauchy_policy}
			\pi_{\bm{\theta}}(a|s) = \frac{1}{\sigma \pi (1+((a-\varphi(s)^\top \bm{\theta})/\sigma)^2) },
		\end{align}
\end{example} 

Interestingly, for the Cauchy distribution (cf. Example \ref{eg:alpha_stable}), the score function is given by 
\begin{align}\label{second_gradient}
\nabla\log \pi_{\bm{\theta}} (s,a)  = \frac{2((a-\varphi(s)^\top\bm{\theta}_{1})/{\sigma})}{1+ {((a-\varphi(s)^\top\bm{\theta}_{1})/{\sigma})^2 }} \left(  \frac{\varphi(s)}{\sigma} \right) ,
\end{align}
for any $\sigma>0$. From \eqref{second_gradient}, we can conclude that $\|\nabla\log \pi_{\bm{\theta}} (s,a)\| \leq \frac{D}{\sigma}$ for all $s$, $a$, and $\bbtheta$. 

With potential choice of policy parameterization detailed, we take a closer look at the relative merits and drawbacks. Intuitively, policies that select actions far from a learned mean parameter over actions may be beneficial in problems where state space coverage is essential. However, employing a Gaussian can result in persistent bias. To see how these issues manifest in practice, we develop the Pathological Mountain Car example next.
\begin{figure}[t]
	\centerline{\includegraphics[scale=0.15]{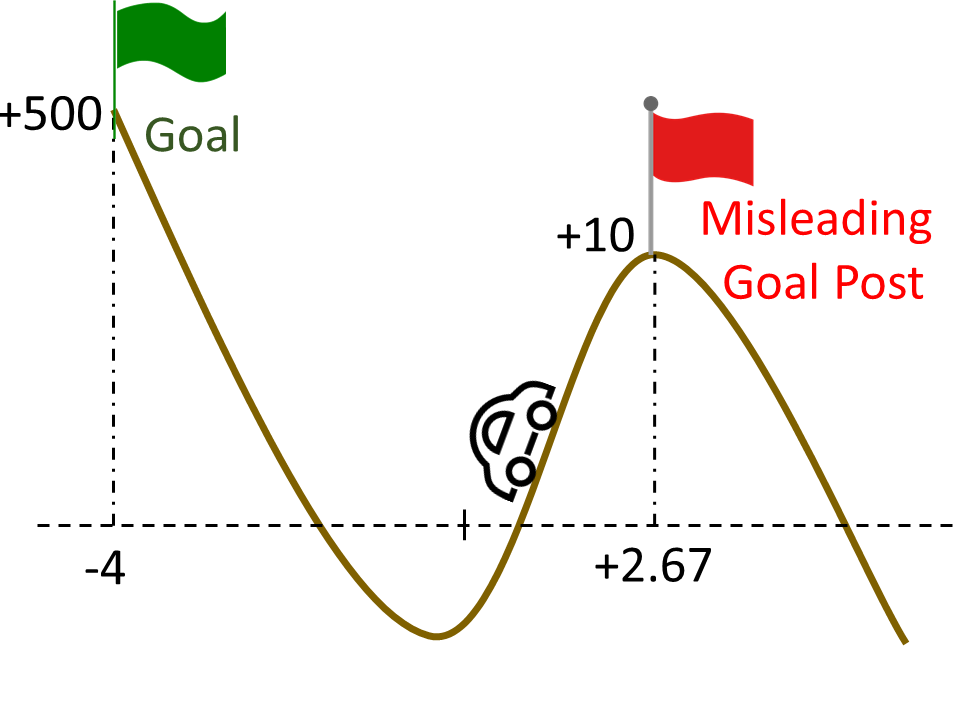}}
	\vspace{0mm}   
	{{\centering  
			\caption{ Pathological Mountain Car example to demonstrate scenario when long and short-term incentives may be misaligned in continuous space: note there is a low reward state (red) and another high reward (red) state atop a higher hill. Policies that do not incentivize exploration get stuck at the spurious goal.} \label{fig:env}}}\vspace{-0mm} 
\end{figure}

{\bf \noindent Pathological Mountain Car.}
Consider an environment with a car trapped between two mountains of different heights as shown in Fig. \ref{fig:env}. Here,  $s\in [-4.0, \,  3.709]$ denotes the state space, and the action $a$ is a one-dimensional scalar representing the speed of the vehicle $\dot{s}$.  The environment consists of two goal posts, a less-rewarding goal (red) at $s$$=$$2.667$ with a reward of $10$ and a high reward (green) at $s$$=$$-4.0$ of $500$. 
 If we consider a Gaussian as a policy for this environment, then $\mathcal{A}$ is $\mathbb{R}$. So there are two issues with using Gaussian policy for this environment.  One major issue is in practice, we always define a set $\mathcal{C}$ which denotes the practical range of actions around mean. For instance, we define $\mathcal{C}:=[-5,5]$ in the experiments (cf. \ref{sec:simulations}) which comes from the practical limitations of the car. It prioritizes the actions around the mean, which can result in becoming stuck at the less rewarding states, never reaching the highest reward. We remark that this is the case with the experiments in existing literature as well (see Table \ref{tab12}). Therefore, there will always be a finite $\lambda$, and hence policy gradients converge to a policy that is not even a stationary point, but instead biased. 

While encouraging action selection far from the mean may seem practically beneficial, and advantageous from the perspective of defining a policy whose score function is bounded, and hence mitigates the hidden bias of \eqref{define_Set},  doing so exhibits a downside. In particular, heavy-tailed policies result in search directions which may be unstable due to the high probability of taking actions far from mean. 
Surmounting this issue is the focus of Sec. \ref{section:algorithm222}, where we develop a mirror ascent-type algorithm for \eqref{eq:main_prob}, discuss existing divergence issues with it due to the interaction of non-convexity and standard parameter selections. Consequently, we put forth a novel momentum variant, which is our main algorithmic innovation.

\begin{algorithm}[t]
	\begin{algorithmic}[1]
		\STATE \textbf{Initialize} :   Initial parameters,  $\bm{\theta}_0$, $\beta$, $\gamma$, step-size $\eta$, $\bbg_0 =0$  \\
		\textbf{Repeat for $k=1,\dots$}
		\STATE Sample trajectories $\xi_k(\bbtheta_{k})$ and $\xi_k(\bbtheta_{k-1})$ of length  $T_k\sim\text{Geom}(1-\gamma^{1/2})$ using policies $\pi_{\bm{\theta}_{k}}$ and $\pi_{\bm{\theta}_{k-1}}$
		\vspace{-0mm}
		\STATE  Estimate $\widetilde{\nabla}  {J}(\bm{\theta}_{k-1}, \xi_k(\bbtheta_{k}))$, ${\nabla} J(\bm{\theta}_k, \xi_k(\bbtheta_{k}))$  via \eqref{eq:policy_gradient_iteration2} ans \eqref{eq:policy_gradient_iteration}, respectively \vspace{-0mm}
		\STATE  {$\hat \bbg_k$=$(1-\beta)(\hat \bbg_{k-1}$-$\widetilde{\nabla} J(\bm{\theta}_{k-1}, \xi_k(\bbtheta_{k})))$+$ {\nabla} J(\bm{\theta}_k, \xi_k(\bbtheta_k))$}\vspace{-0mm}
		\STATE $\bm{\theta}_{k+1} = \argmax_{\bm{\theta} } \{\langle \hat \bbg_k,\bm{\theta}\rangle-\frac{1}{\eta}D_{\psi}(\bm{\theta},\bm{\theta}_k)\}$
		\STATE $k \leftarrow k+1$
		
		\textbf{ Until Convergence}
		\STATE \textbf{Return: $\bm{\theta}_k$} \\ 
	\end{algorithmic}
	\caption{Policy Gradient with Stochastic Recursive Mirror Ascent (SRMA)}
	\label{Algorithm2}
\end{algorithm}
\section{Stochastic Recursive Mirror Ascent}\label{section:algorithm222}
To conduct policy search when employing heavy-tailed parameterizations such as the Cauchy, we note that extreme action selection can cause numerical instabilities in the sequence of policy parameters in practice. To mitigate this issue, we take inspiration from proximal policy optimization \citep{schulman2017proximal}, which restricts movement of policy parameters through
 regularization of the gradient update. In the stochastic setting, such regularization can be rigorously substantiated through stochastic mirror ascent \citep{nemirovski2009robust,ghadimi2016mini,yang2019policy}. Next, we discuss the technical limitations of existing stochastic mirror ascent approaches, which necessitate a modification that uses an additional recursive averaging step. The  stochastic mirror ascent (SMA) update for \eqref{eq:main_prob} is given by
\begin{align}\label{optimization_problem2}
	\bm{\theta}_{k+1} =\argmax_{\bm{\theta} } \Big\{\big\langle \bbg_k ,\bm{\theta}\big\rangle-\frac{1}{\eta}D_{\psi}(\bm{\theta},\bm{\theta}_k)\Big\},
\end{align}
where $\bbg_k:={\nabla} J(\bm{\theta}_k, \xi_k(\bbtheta_k))$ and $D_{\psi}$ denotes a Bregman divergence defined with respect to the strongly convex function $\psi(\bbx)$ with $\zeta$ as the strong convexity parameter. We remark that the update in \eqref{optimization_problem2} boils down the the standard stochastic gradient ascent (policy gradient in \eqref{eq:policy_gradient_iteration2}) for $\psi(\bbtheta)=\frac{1}{2}\|\bbtheta\|^2$, and when $\psi(\bbtheta) = \sum_j [\bbtheta]_j   \log [\bbtheta]_j$ is the KL divergence, and the policy is tabular $\pi_{\bbtheta}=\pi$, this reduces to Natural Policy Gradient. Extensions to parameterized settings are possible with Fischer information approximations, but we do not discuss this further. To analyze the update in \eqref{optimization_problem2}, we define the Bregman gradient 
\begin{equation}\label{eq:bregman_grad}
\mathcal{G}_{\eta,\bbg_k}^\psi(\bm{\theta}_{k})\!=\!\big(\theta_k\!-\!\argmax_{\bm{\theta}} \Big\{\big\langle \bbg_k ,\bm{\theta}\big\rangle-\frac{1}{\eta}D_{\psi}(\bm{\theta},\bm{\theta}_k)\Big\} \big)/\eta
\end{equation}
 corresponding to the stochastic estimate of the gradient $\hat{\nabla} {J}(\bbtheta_k) $ as an analogue of the fixed point of a gradient projection update, but instead with respect to the Bregman divergence --  see \citep{ghadimi2016mini,yang2019policy}. Doing so allows us to rewrite  \eqref{optimization_problem2} as 
\begin{align}\label{modified}
	\bm{\theta}_{k+1}  = \bm{\theta}_{k}+ \eta  \mathcal{G}_{\eta,\bbg_k}^\psi(\bm{\theta}_{k}).
\end{align} 

\noindent \textbf{Optimality Criteria.}  To characterize convergence to first-order stationarity, we analyze the attenuation rate of the norm of the generalized gradient to a small constant $\epsilon$ as  $\mathbb{E}\left[\|\mathcal{G}_{\eta,\bbg_k}^\psi(\bbtheta)\|^2\right]\leq\epsilon$ which defines $\epsilon$ first-order stationarity, which is standard in the analysis of mirror ascent \citep{ghadimi2016mini}. Importantly, in \citep{ghadimi2016mini}, for batch size (number of stochastic gradient samples per iteration) of $B_k=\mathcal{O}(1)$, for any non-convex stochastic programming problem, under suitable conditions, we have that
\begin{align}\label{lower_bound}
	\mathbb{E}\left[\|\mathcal{G}_{\eta,\bbg_k}^\psi(\bbtheta)\|^2\right]\leq \frac{D_{\psi}+\frac{\sigma^2}{\zeta}\sum_{k=1}^{K}\eta_k}{\sum_{k=1}^{K}(\zeta\eta_k-L\eta_k^2)},
\end{align}
where $\eta_k$ is the step size used at each $k$. From the right hand side of \eqref{lower_bound}, we can conclude that it is lower bounded by $\frac{\sigma^2}{\zeta^2}$ independent of choice of step-size. To address this issue, one must allow the batch size $B_k$ to increase with $k$ to obtain convergence \citep{ghadimi2016mini,yang2019policy}. But obtaining a convergent algorithm for SMA algorithm with general non-convex \emph{un-regularized} objective and fixed batch size per iteration remains a challenge. Moreover, increasing batch sizes at an unbounded rate is impractical in RL, as it is quite costly to sample ever-greater numbers of trajectories in between policy updates.

In this work, we address this issue via a recursive averaging step together with a difference of two gradient evaluations, i.e., gradient tracking, at each $k$ given as
\begin{align}
	\hat \bbg_k=&\left(1-\beta\right)(\hat \bbg_{k-1}\!-\!{\nabla} J(\bm{\theta}_{k-1}, \xi_k(\bm{\theta}_{k})))\label{gradient_Estimate00}
	\\ &\hspace{4cm} + {\nabla} J(\bm{\theta}_k, \xi_k(\bm{\theta}_{k}))\; , 	\nonumber 
	\\
	\bm{\theta}_{k+1} =&\argmax_{\bm{\theta} } \{\langle \hat \bbg_k,\bm{\theta}\rangle-\frac{1}{\eta}D_{\psi}(\bm{\theta},\bm{\theta}_k)\},	\label{proximal_update00} \
\end{align}
 We introduce $\beta$ as the step size for the gradient tracking update in \eqref{gradient_Estimate00}. Different from the standard SMA update in \eqref{optimization_problem2}, at each $k$, we need access to the stochastic gradients ${\nabla} J(\bm{\theta}_k, \xi_k(\bbtheta_k))$ and ${\nabla} J(\bm{\theta}_{k-1}, \xi_k(\bbtheta_{k}))$ mentioned in \eqref{gradient_Estimate00} evaluated at two different instances $\bbtheta_{k-1}$ and $\bbtheta_{k}$. 
 %
%
%
Here, $\xi_k(\bbtheta_k)$ denotes the state action pair trajectory $\{(s_0,a_0,r_0)\cdots(s_{T_{k}},a_{T_{k}},r_{T_{k}})~|~ a_k\sim \pi_{\bbtheta_k}(\cdot~|~ s_k)\}$ sampled using policy $\pi_{\bbtheta_k}$. 
%
%
%
%
The update in \eqref{gradient_Estimate00} is  motivated from the momentum based tracking update proposed in \cite{cutkosky2019momentum}, where two stochastic gradient evaluated using two different  parameter instances at the same random variable are utilized for the gradient tracking. But using \eqref{gradient_Estimate00} in its current form results in a well known distribution shift issue (also appears in \cite{papini2018stochastic,yuan2020stochastic}) because the trajectory sampling distribution (which we show explicitly by writing $\xi_k(\bbtheta_k)$ as function of $\bbtheta_k$) now depends upon parameter $\bbtheta$ which is not the case in \cite{cutkosky2019momentum}. To resolve this issue, we employ an importance sampling (IS) based-modifications of the updates in \eqref{modified22}-\eqref{proximal_update} given as
\begin{align}
	\hat \bbg_k=&\left(1-\beta\right)(\hat \bbg_{k-1}\!-\!\widetilde{\nabla} {J} (\bm{\theta}_{k-1}, \xi_k(\bbtheta_{k}))) \label{modified22}
	\\
	&\hspace{3cm}+ {\nabla} J(\bm{\theta}_k, \xi_k(\bbtheta_k))\; \nonumber
	\\
	\bm{\theta}_{k+1} =&\argmax_{\bm{\theta} } \{\langle \hat \bbg_k,\bm{\theta}\rangle-\frac{1}{\eta}D_{\psi}(\bm{\theta},\bm{\theta}_k)\},	\label{proximal_update} \
\end{align} 
where we define 
\begin{align}\label{eq:policy_gradient_iteration2}
	\widetilde{\nabla} & {J}(\bm{\theta}_{k-1}, \xi_k(\bbtheta_{k}))\\
	=& \sum_{t=0}^{T_k}\gamma^{t/2}r_t\cdot w_t(\tau|\bbtheta_{k-1},\bbtheta_{k})\cdot\sum_{\tau=0}^{t}\nabla\log\pi_{\bm{\theta}_{k-1}}(a_{\tau}\given s_{\tau}),\nonumber
\end{align}
%
where $w_t(\tau|\bbtheta_{k-1},\bbtheta_{k})$ is the importance sampling weight given by $w_t(\tau|\bbtheta_{k-1},\bbtheta_{k})=\frac{\prod_{h=0}^t \pi_{\bbtheta_{k-1}(a_h,s_h)}}{\prod_{h=0}^t \pi_{\bbtheta_{k}(a_h,s_h)}}.$ We call the updates in \eqref{modified22}-\eqref{proximal_update} as  Stochastic Recursive Mirror Ascent (SRMA). The overall proposed scheme for RL is summarized as Algorithm \ref{Algorithm2}. Its important to note that this instantiation of gradient tracking is novel, in that previous uses were explicitly for variance reduction of non-proximal (Euclidean updates), whereas here they are specifically employed to reduce persistent bias [cf. \eqref{lower_bound}] associated with stochastic mirror ascent with constant batch size.
Next, we shift towards establishing convergence of Algorithm \ref{Algorithm2}. 

\section{Convergence Analysis} \label{sec:convergence}

%
%
%
In this section, we establish that Algorithm \ref{Algorithm2} converges to stationarity in the sense of $\mathbb{E}\left[\|\mathcal{G}_{\eta,\bbg_k}^\psi(\bbtheta)\|^2\right]\leq\epsilon$, with $\mathcal{G}_{\eta,\bbg_k}^\psi(\bbtheta)$ as in \eqref{eq:bregman_grad}. 
Without loss of generality, we reformulate the problem in the syntax of minimization, that is, we consider minimizing a function $F(\bbtheta):=-J(\bbtheta)$, with $J(\bbtheta)$ as the cumulative return under policy $\pi_{\bbtheta}$ in \eqref{eq:main_prob}. Hence, the problem we consider for the analysis is given by
\begin{align}\label{modified_problem}
	\min_{\bbtheta} F(\bbtheta).
\end{align}
Let $\nabla F(\bbtheta_k)$ denote the gradient of function $F(\bbtheta)$ at $\bbtheta_k$ and $\xi_k(\bbtheta_k)$ is the randomness introduced at $k$ to estimate the policy gradient. The associated stochastic unbiased gradient estimate is denoted as  $\nabla F(\bbtheta_k,\xi_k(\bbtheta_k))$. 
%
With this modified expression for the gradient and the reformulation in terms of minimization \eqref{modified_problem}, the parameter update for $\bbtheta_{k}$ may be rewritten as
\begin{align}
 &\hat \bbg_k\!=\!(1\!-\!\beta)(\hat \bbg_{k-1}\!-\!\widetilde{\nabla} F(\bm{\theta}_{k-1}, \xi_k(\bbtheta_{k})))\!\!+\!\! {\nabla} F(\bm{\theta}_k, \xi_k(\bbtheta_k)), \nonumber
	\\
&	\bm{\theta}_{k+1} =\argmin_{\bm{\theta} } \big\{\langle \hat \bbg_k,\bm{\theta}\rangle+\frac{1}{\eta}D_{\psi}(\bm{\theta},\bm{\theta}_k)\big\}.	\label{gradient_Estimate12}
\end{align}
Next, we present the convergence rate results for the updates in \eqref{gradient_Estimate12} for any general non-convex function $F(\bbtheta)$ which may be of independent interest.  First, let us define the filtration  
	$\mathcal{F}_{k}:=\big\{\xi_u\big\}_{u<k},$
which collects randomness associated with trajectory realizations until step $k$. We use the definition of $\mathcal{F}_{k}$ to state the following assumptions required for the following analysis. 
\begin{assumption}(Boundedness)\label{assum:reward}
	The reward function $r(s,a)$ is bounded as $|r(s,a)|\leq U_R$ for all $s$ and $a$. The state features $\varphi(s)\in\mathbb{R}^d$, are bounded, i.e., for some $D>0$,  $\|\varphi(s)\|\leq D$.
\end{assumption}
\begin{assumption}(IS Variance)\label{importance_variance}
	For any policy parameters $\bbtheta_1$ and $\bbtheta_2$, let the importance sampling weights defined as $w(\tau|\bbtheta_1,\bbtheta_2):= \frac{p(\tau~|~\bbtheta_{1})}{p(\tau~|~\bbtheta_{2})}$. Then it holds that $Var\left(w(\tau|\bbtheta_1,\bbtheta_2)\right)$$\leq$$ C$, for some $C$$>$$0$ where $\tau\sim p(\cdot|\bbtheta_2)$. 
\end{assumption}
 Note that when the non-convex objective $F(\bbtheta)=-J(\bbtheta)$, then  in the policy search setting, then above assumptions are sufficient for the SRMA algorithm to converge to the exact stationary point. But since, we are interested in doing the convergence analysis for general non-convex objective $F(\bbtheta)$, we mention next the general set of Assumptions \ref{assum:first1}-\ref{ass:smooth} required for $F(\bbtheta)$. For the RL setting considered in this paper where $F(\bbtheta)=-J(\bbtheta)$, we prove in  Appendix \ref{assumptions_Satisfied} that all the Assumptions \ref{assum:first1}-\ref{ass:smooth} are satisfied. 
\begin{assumption}\label{assum:first1}
	The stochastic estimate \eqref{eq:policy_gradient_iteration} is unbiased: $\mathbb{E}[\nabla F(\bbtheta_k,\xi_k)~|~\mathcal{F}_k]= \nabla F(\bbtheta_k)$ for all $k$. 	
\end{assumption}
\begin{assumption}\label{assum:second1}
	The variance of the stochastic gradient satisfies the growth condition  $$ \mathbb{E}\left[\|\nabla F(\bbtheta_k,\xi_k)- \nabla F(\bbtheta_k)\|^2\right]\leq {m_0 + m_1\|\nabla F(\bbtheta_k)\|^2} $$ for all $k$ where $m_0>0$ and $m_1\geq 0$ are scalars. 
\end{assumption}
\begin{assumption}\label{assum:third1}
	{The original gradient $\nabla F(\bbtheta_k)$ and the stochastic Bergman gradient $\mathcal{G}_{\eta,\hbg_k}^\psi(\bbtheta_k)$ satisfy the variance growth condition
		%
			$$\mathbb{E}\left[\|\nabla F(\bbtheta_k)\!-\! \mathcal{G}_{\eta,\hbg_k}^\psi(\bbtheta_k)\|^2\right]\leq m_2+m_3\mathbb{E}\left[\|\mathcal{G}_{\eta,\hbg_k}^\psi(\bbtheta_k)\|^2\right]$$
		where $m_2>0,m_3\geq0$ are scalars, and $\hbg_k$ is as in \eqref{gradient_Estimate00}. }
\end{assumption}
\begin{assumption}\label{ass:smooth}
The objective $F(\cdot)$ is $L$-smooth. 
\end{assumption}
%
Assumptions \ref{assum:first1}-\ref{ass:smooth} are standard in the optimization literature.  Assumption \ref{assum:third1} imposes a bound between the original gradient estimate and the generalized gradient evaluated at the current biased estimate of the gradient $\hbg_k$. Observe for the case when $\psi(\bbx)=\frac{1}{2}\|\bbx\|^2$ and we utilize the stochastic unbiased gradient at each $k$, Assumption \ref{assum:third1} simplifies to Assumption \ref{assum:second1}. Assumption \ref{ass:smooth} is related to the smoothness of the objective function $F$.

Before proceeding with the main result of this work, we recall two important properties of the generalized gradient from \citep[Lemma 1]{ghadimi2013stochastic}:
\begin{align}\label{prop1}
	\langle \bbg_k,\mathcal{G}_{\eta,\bbg_k}^\psi(\bm{\theta}_{k})\rangle \geq \zeta \|\mathcal{G}_{\eta,\bbg_k}^\psi(\bm{\theta}_{k})\|^2 \\
%
%
\label{prop2}
	\|\mathcal{G}_{\eta,\bbg_1}^\psi(\bm{\theta}) -\mathcal{G}_{\eta,\bbg_2}^\psi(\bm{\theta})\| \leq \frac{1}{\zeta}\|\bbg_1-\bbg_2\|.
\end{align}

The inequalities \eqref{prop1} and \eqref{prop2} will be used in the analysis. Next, we present an intermediate lemmas which bounds the stochastic errors associated with gradient estimation $\mathbb{E}\left[\|\hbg_k-\nabla F(\bbtheta_k)\|^2\right]$. 
\begin{lemma}\label{cauchy_bounded}
	The norm of the score function is bounded as $\| \nabla \log \pi_{\bm{\theta}} (a|s)\|\leq B$ for all $s\in\mathcal{S}$ and $a\in\mathcal{A}$ for a Cauchy policy.
\end{lemma}
See Appendix \ref{bounded} for the proof of Lemma \ref{cauchy_bounded}. We note that the statement of Lemma \ref{cauchy_bounded} explicitly provide a bound on the norm of score function which is usually assumed in the existing PG methods \cite{zhang2020global,yuan2020stochastic,xu2017stochastic,xu2019sample}. However, it cannot be assumed due to its severe dependence on the choice of policy parameterization. Next, we move to bound the error between the gradient $\hbg_k$ used in Algorithm \ref{Algorithm2} and the original gradient $\nabla F(\bbtheta_k)$ as follows. 
	\begin{lemma}\label{lemtrack}
	Let $\e_k := \mathbb{E}\left[\|\hbg_k-\nabla F(\bbtheta_k)\|^2\right]$, then for all $k\geq 1$, it holds that
	\begin{align}
		\e_{k} \leq & (1-\beta)^2\e_{k-1} + 2\eta^2 {L_{1}}{\mathbb{E}\left[\|\mathcal{G}_{\eta,\hbg_k}^\psi(\bbtheta_k)\|^2\right]} + 2m_0\beta^2\nonumber
		\\
		&{+2m_1\beta^2\|\nabla F(\bbtheta_{k})\|}^2 \label{track},
	\end{align}
where $L_1:=\left(2L^2+\frac{2C_wU_R^2B^2(1+\sqrt{\gamma})}{(1-\gamma)((1-\sqrt{\gamma})^2)}\right)$.
\end{lemma}
See Appendix \ref{proof_lemma} in the Appendix for proof. The result in Lemma \ref{lemtrack} bounds the per step expected value of the the norm of error for each $k$.  Next we present the main theorem of this paper. We note that the tracking result in Lemma \ref{lemtrack} incorporates random horizon $T_k$ into derivation which is missing in the existing literature \cite{papini2018stochastic,yuan2020stochastic,xu2017stochastic}. 
\begin{theorem}\label{main_khorem}
Under Assumption \ref{assum:first1}-\ref{ass:smooth}, with step-size selections $\beta$$=$$C_1\eta$ with $C_1> 0$ and $\eta\leq  \min\Big\{\frac{\zeta L'}{10},\frac{\zeta}{8\tilde{m}_3 C_1^2}\Big\}$, in order to achieve 
%
		$\min_{1\leq k\leq K}\mathbb{E}\left[\|\mathcal{G}_{\eta,\hbg_k}^\psi(\bbtheta_k)\|^2\right]\leq \epsilon $
%
with $\epsilon \leq \min\Big\{\frac{\zeta L'}{10},\frac{\zeta}{8\tilde{m}_3 C_1^2}\Big\}$, the iterates in Algorithm \ref{Algorithm2} requires at least $K\geq \mathcal{O}\left(\frac{1}{\epsilon^2}\right)$ iterations with $\mathcal{O}(1)$ stochastic gradients samples at each $k$. Additionally, in order to achieve $		\min_{1\leq k\leq K}\mathbb{E}\left[\|\mathcal{G}_{\eta,\hbg_k}^\psi(\bbtheta_k)\|\right]\leq \epsilon$, the iterates in Algorithm \ref{Algorithm2} requires at least $K\geq \mathcal{O}\left(\frac{1}{\epsilon^4}\right)$ with $\mathcal{O}(1)$ stochastic gradients samples at each $k$.
\end{theorem}
%
 We note that a related but simpler specification of step-size $\eta$ also permitted: $\eta=\frac{\eta_0}{\sqrt{K}}$ in terms of final iteration index $K$ with $\eta_0=  \min\Big\{\frac{\zeta L'}{10},\frac{\zeta}{8\tilde{m}_3 C_1^2}\Big\}$. 
%
See Appendix \ref{proof_theorem} for proof. Note that the use of recursive update for the stochastic gradient estimate in \eqref{modified22} permits us to achieve the $\mathcal{O}(\frac{1}{\epsilon^2})$ oracle complexity with a 
\begin{table}[t]
	\centering
	\resizebox{0.9\columnwidth}{!}{
		\begin{tabular}{|l|l|l|}
			\hline
			Refs.& Samples per Iteration & Sample Complexity \\
			\hline
			\citep{ghadimi2016mini} & $\mathcal{O}({1}/{\epsilon})$& $\mathcal{O}({1}/{\epsilon^2})$ \\
			%
			%
			%
			\citep{yang2019policy} & $\mathcal{O}({1}/{\epsilon})$& $\mathcal{O}({1}/{\epsilon^2})$ \\
			%
			%
			%
			This work& $\mathcal{O}(1)$& $\mathcal{O}({1}/{\epsilon^2})$ \\
			\hline
		\end{tabular}
	}
	\caption{Summary of related results in optimization literature to achieve $\mathbb{E}\left[\|\mathcal{G}_{\eta,\hbg_k}^\psi(\bbtheta_k)\|^2\right]\leq \epsilon $ .} 
	\label{tab12}
	\vspace{-0mm}
\end{table}
\begin{figure*}[ht]
	\centering
	\subfigure[]
	{\includegraphics[width=.5\columnwidth,clip = true]{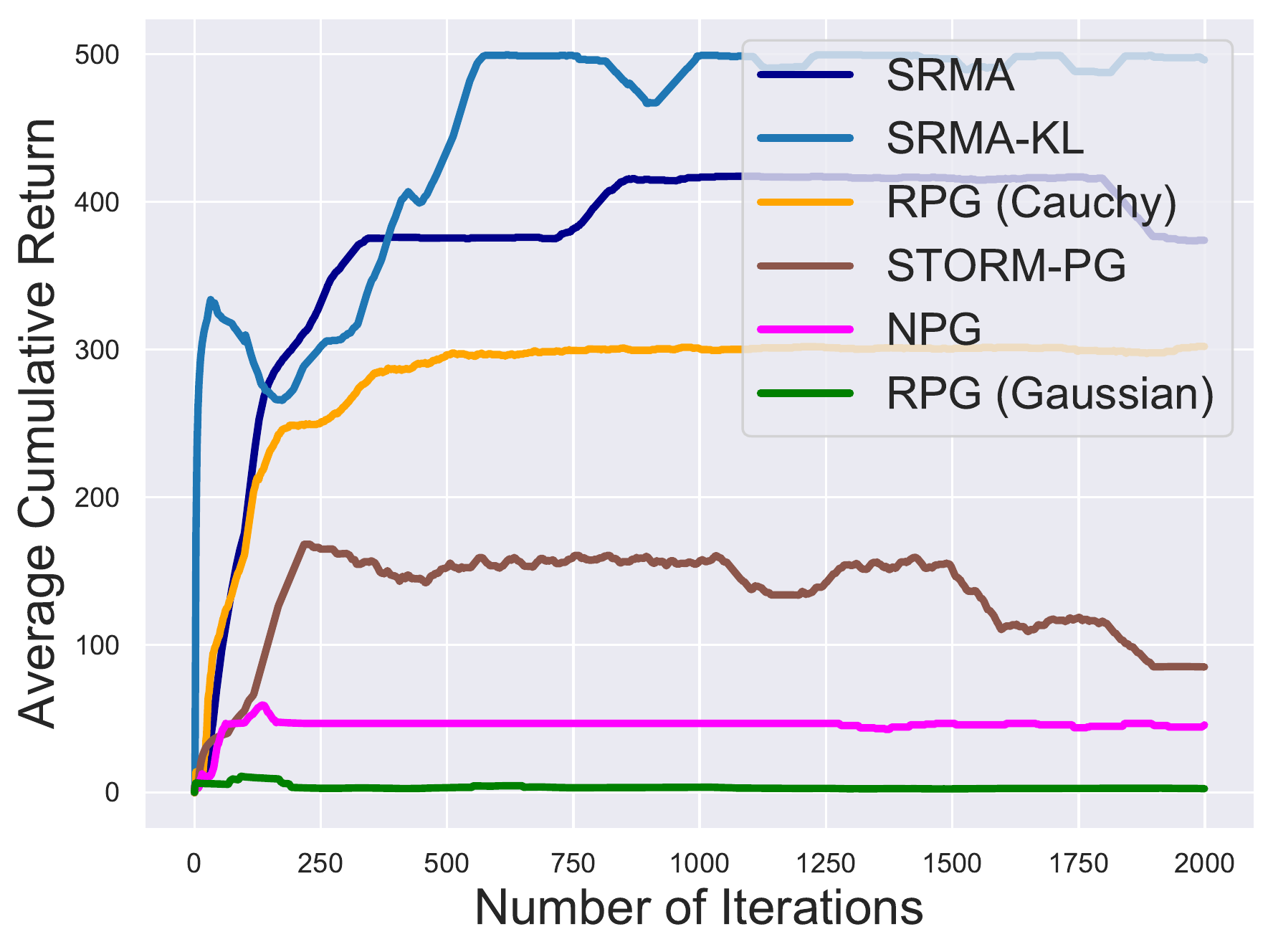} \label{fig:1}}
	\hspace{0mm}   
	\subfigure[]
	{\includegraphics[width=.5\columnwidth,clip = true]{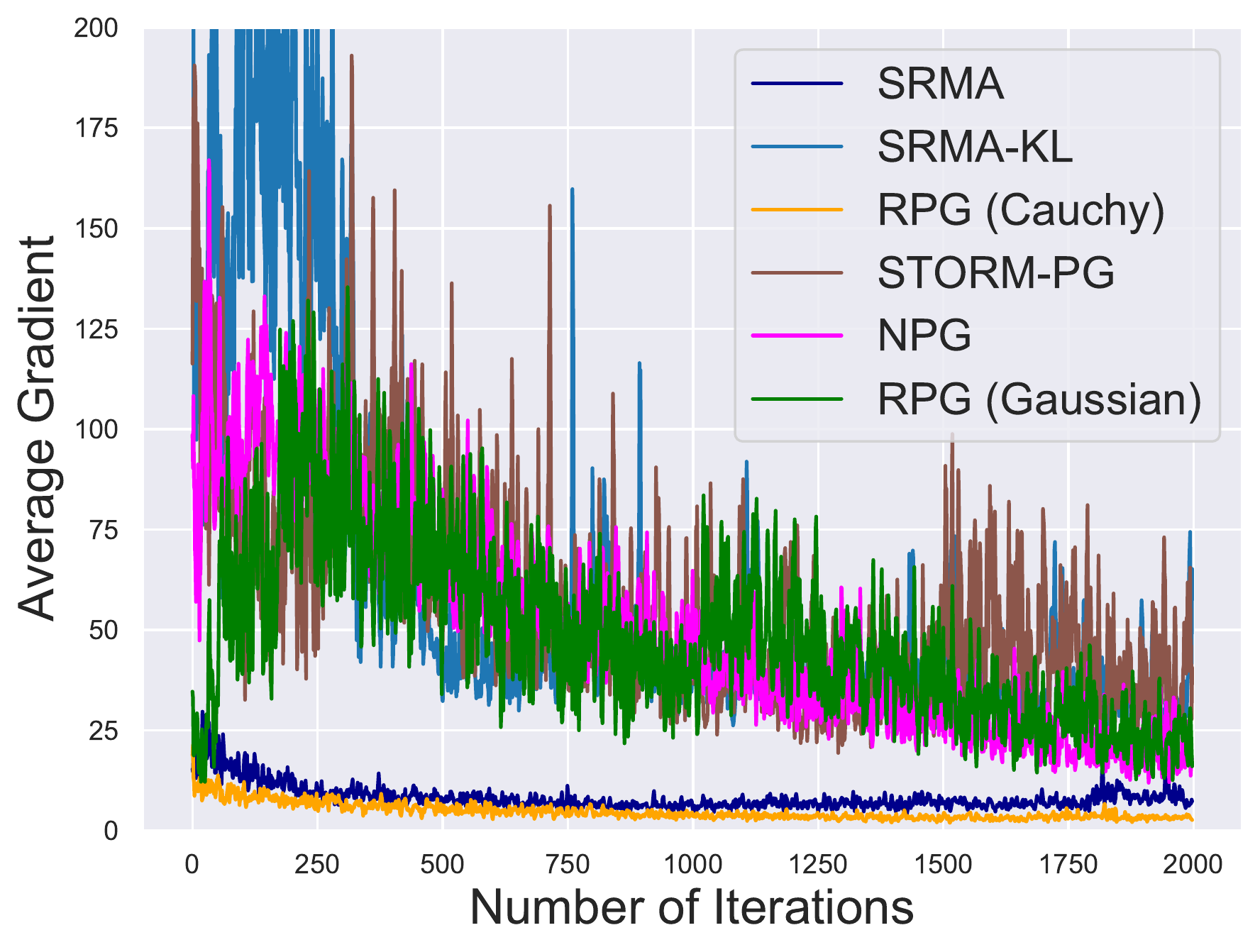}\label{fig:2}}
	\subfigure[]
	{\includegraphics[width=.5\columnwidth,clip = true]{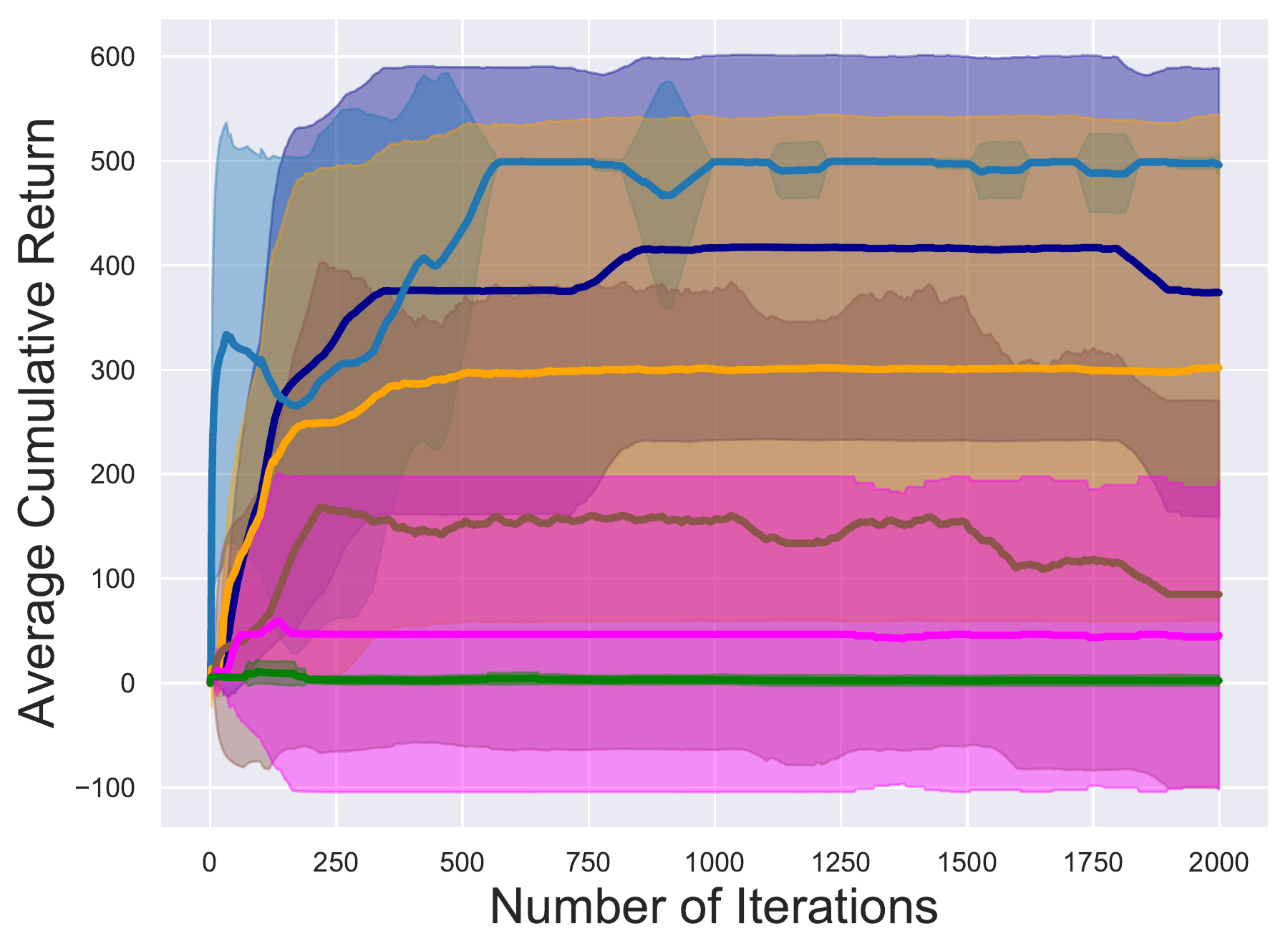}\label{fig:3}}
	\hspace{-0mm}   
		\subfigure[]
		{\includegraphics[width=.5\columnwidth,clip = true]{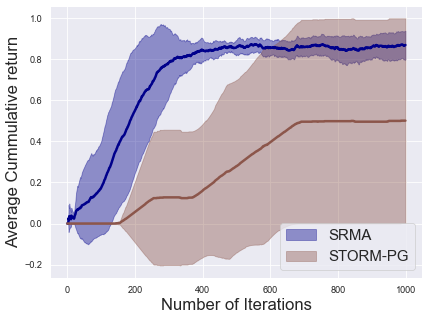} \label{fig:4}}	
	\caption{{\textbf{(a)}} This figure compares the average cumulative return obtained by the proposed  Stochastic Recursive Mirror Ascent (SRMA) algorithm with the existing state of the art methods available to solve MDP in continuous state action space. We considered Pathological Mountain Car environment (cf. Fig. \ref{fig:env}) for the experiments. We note that SRMA-KL is able to achieve the maximum reward as compared to other methods. 
		{\textbf{(b)}} In this figure, we plot the average Bregman gradient (performance metric in Theorem \ref{main_khorem}) for the proposed algorithm and compares with others. We note that even if the gradient norm converges to zero for STORM-PG (which used Gaussian policy parametrization), the corresponding reward is low. This implies that Gaussian policy parametrization is not able to sufficiently explore the environment. 
		{\textbf{(c)}} In this figure, we plot the average cumulative gradient shown  in Fig. \ref{fig:1} but with confidence intervals.  {\textbf{(d)}} Performance comparison of the proposed algorithm with STORM-PG on Mario environment mentioned in \cite{matheron2019problem}.  
		\vspace{-3mm}
	}	
	%
	\label{figure_results}
\end{figure*} 
We remark that the sample complexity result in Theorem \ref{main_khorem} is general enough and holds for any non-convex objective $F(\bbtheta)$ as well. constant batch size of gradients $\mathcal{O}(1)$ per iteration.    In addition, we note that in the stochastic optimization literature, the 
related work on stochastic mirror ascent for  non-convex objectives require an increasing batch size for convergence \citep{ghadimi2013stochastic,yang2019policy}. In contrast, the proposed algorithm is able to achieve the optimal convergence rate with finite batch size which is of great practical importance.  We summarize the results in Table \ref{tab12}. Under RL setting considered in this paper, we prove in Appendix \ref{assumptions_Satisfied} that all the assumptions are satisfied. Further, in the PG analysis in the existing literature, $\mathbb{E}\left[\|\mathcal{G}_{\eta,\hbg_k}^\psi(\bbtheta_k)\|\right]\leq \epsilon$ is used as a performance metric, which requires $\mathcal{O}\left(\frac{1}{\epsilon^4}\right)$ (as mentioned in Table \ref{literature}) sample complexity with batch size of $\mathcal{O}(1)$. Additionally, employing the Cauchy policy parametrization removes the bias present in the convergence of existing PG methods in continuous spaces \cite{bedi2021sample}. In other words, $\lambda$ (cf. \eqref{define_Set}) is exactly zero for the proposed algorithm, which results in exact convergence to stationarity.
%
%
 We evaluate the experimental utility of SRMA next.
 %



\section{Experiments} \label{sec:simulations}

This section validates the efficacy of proposed heavy tailed (Cauchy distribution) policy parameterizations  in the continuous MDP of the Pathological Mountain Car (PMC) from Section \ref{sec:prob}. We consider an incentive structure in which the amount of energy expenditure, i.e., the action squared, is negatively penalized such that $r(s,a)=-  a^2$ \text{for} $-4.0 < s < 3.7,\,  s \neq 2.6$, $r(s,a)=500 -  a^2$ if $s = -4.0$, and $r(s,a)=10-   a^2$ if $s = 2.6$.
%
At each episode, the position of the car is initialized uniformly at random from $[1.15,2.0]$.   
State $s$ is constrained  to an interval  $[-4.0, \,  3.7]$ and action $a$ lies in $[-6, \, 6]$. The discounted factor $\gamma$ is  $0.97$ and we use a step size of $0.005$.
In Fig.~\ref{figure_results}, we  compare  performance of the proposed SRMA algorithm with the other existing state of the art techniques for continuous MDP such as NPG \cite{kakade2002natural}, RPG \cite{zhang2020global}, STORM-PG \cite{yuan2020stochastic}. For experiments, we implemented two versions of the proposed SRMA algorithm called SRMA and SRMA-KL (see legends in Fig~\ref{fig:1}). For the SRMA-KL version, we use KL divergence between the policies at $k$ and $k-1$ as the Bregman divergence term, which is also used for NPG implementation. In Fig~\ref{fig:1}, SRMA denotes implementation where we used standard Euclidean distance as the Bregman term. For SRMA and SRMA-KL, we used Cauchy policy parametrization, and for RPG (Gaussian), NPG, and STORM-PG, we have used Gaussian policy parametrization.
%


Fig.~\ref{fig:1} shows the cumulative  return averaged over $15$ runs of the experiments. It is clear that the SRMA algorithm which utilizes Cauchy policy parametrization yields better performance as relative to comparators which use Gaussian parametrization.  Moreover, we note that simply replacing Gaussian parametrization by a Cauchy results in numerical instability, as may be observed by RPG (Cauchy), which is PG iteration with a Cauchy policy. This instability is corroborated in Fig. \ref{fig:3} where we observe the average rewards return along with standard deviation across runs, which is large for the orange curve. This justifies the use of policy mirror ascent-type updates. We also plot the average Bregman gradient in Fig.~\ref{fig:2} to show the convergence of the proposed algorithm under constant step-size and batch-size as compared with the others methods available in literature. 
\section{Conclusion} \label{sec:conclusion}
In this work, we focused on policy gradient method for solving RL problems associated with infinite-horizon discounted returns. In some problems, the one-step reward may be very far from the value of a given state, which can cause policies to become mired at spurious behavior. Inspired by the relationship between score function boundedness and persistent bias that may arise when operating in continuous action spaces, we proposed to study policy search under heavy-tailed parameterizations. This parameterization introduced numerical challenges, namely, numerical instability in the agent's trajectory, and  potentially volatile changes in policy gradients. 
To address this issue, we studied mirror ascent-type updates, which we stabilized with recursive gradient tracking. The convergence of the resulting iterative schemes was established under novel error bound conditions for the generalized Bregman gradient. Moreover, experimentally, we observed favorable performance of the proposed approach for escaping spurious stationary points. 
 
%
\bibliography{bibliography}

\begin{thebibliography}{50}
\providecommand{\natexlab}[1]{#1}
\providecommand{\url}[1]{\texttt{#1}}
\expandafter\ifx\csname urlstyle\endcsname\relax
  \providecommand{\doi}[1]{doi: #1}\else
  \providecommand{\doi}{doi: \begingroup \urlstyle{rm}\Url}\fi

\bibitem[Agarwal et~al.(2020)Agarwal, Kakade, Lee, and
  Mahajan]{agarwal2020optimality}
Agarwal, A., Kakade, S.~M., Lee, J.~D., and Mahajan, G.
\newblock Optimality and approximation with policy gradient methods in markov
  decision processes.
\newblock In \emph{Conference on Learning Theory}, pp.\  64--66. PMLR, 2020.

\bibitem[Barab{\'a}si et~al.(2003)]{barabasi2003emergence}
Barab{\'a}si, A.-L. et~al.
\newblock Emergence of scaling in complex networks.
\newblock \emph{Handbook of Graphs and Networks: From the Genome to the
  Internet. Berlin: Wiley-VCH}, 2003.

\bibitem[Baxter \& Bartlett(2001)Baxter and Bartlett]{baxter2001infinite}
Baxter, J. and Bartlett, P.~L.
\newblock Infinite-horizon policy-gradient estimation.
\newblock \emph{Journal of Artificial Intelligence Research}, 15:\penalty0
  319--350, 2001.

\bibitem[Bedi et~al.(2021)Bedi, Parayil, Zhang, Wang, and
  Koppel]{bedi2021sample}
Bedi, A.~S., Parayil, A., Zhang, J., Wang, M., and Koppel, A.
\newblock On the sample complexity and metastability of heavy-tailed policy
  search in continuous control.
\newblock \emph{arXiv preprint arXiv:2106.08414}, 2021.

\bibitem[Bellman(1957)]{bellman57a}
Bellman, R.~E.
\newblock \emph{Dynamic Programming}.
\newblock Courier Dover, 1957.
\newblock ISBN 0486428095.

\bibitem[Bertsekas \& Shreve(2004)Bertsekas and
  Shreve]{bertsekas2004stochastic}
Bertsekas, D.~P. and Shreve, S.
\newblock \emph{Stochastic optimal control: the discrete-time case}.
\newblock 2004.

\bibitem[Bhandari \& Russo(2019)Bhandari and Russo]{bhandari2019global}
Bhandari, J. and Russo, D.
\newblock Global optimality guarantees for policy gradient methods.
\newblock \emph{arXiv preprint arXiv:1906.01786}, 2019.

\bibitem[Bhatnagar et~al.(2009)Bhatnagar, Sutton, Ghavamzadeh, and
  Lee]{bhatnagar2009natural}
Bhatnagar, S., Sutton, R., Ghavamzadeh, M., and Lee, M.
\newblock Natural actor-critic algorithms.
\newblock \emph{Automatica}, 45\penalty0 (11):\penalty0 2471--2482, 2009.

\bibitem[Bhatt et~al.(2019)Bhatt, Koppel, and Krishnamurthy]{bhatt2019policy}
Bhatt, S., Koppel, A., and Krishnamurthy, V.
\newblock Policy gradient using weak derivatives for reinforcement learning.
\newblock In \emph{2019 IEEE 58th Conference on Decision and Control (CDC)},
  pp.\  5531--5537. IEEE, 2019.

\bibitem[Borkar(2008)]{borkar2008stochastic}
Borkar, V.~S.
\newblock \emph{Stochastic approximation: \relax{A} dynamical systems
  viewpoint}.
\newblock Cambridge University Press, 2008.

\bibitem[Chou et~al.(2017)Chou, Maturana, and Scherer]{chou2017improving}
Chou, P.-W., Maturana, D., and Scherer, S.
\newblock Improving stochastic policy gradients in continuous control with deep
  reinforcement learning using the beta distribution.
\newblock In \emph{International conference on machine learning}, pp.\
  834--843. PMLR, 2017.

\bibitem[Cutkosky \& Orabona(2019)Cutkosky and Orabona]{cutkosky2019momentum}
Cutkosky, A. and Orabona, F.
\newblock Momentum-based variance reduction in non-convex sgd.
\newblock \emph{arXiv preprint arXiv:1905.10018}, 2019.

\bibitem[Devraj \& Meyn(2017)Devraj and Meyn]{devraj2017zap}
Devraj, A.~M. and Meyn, S.~P.
\newblock Zap q-learning.
\newblock In \emph{Proceedings of the 31st International Conference on Neural
  Information Processing Systems}, pp.\  2232--2241, 2017.

\bibitem[Even-Dar et~al.(2003)Even-Dar, Mansour, and
  Bartlett]{even2003learning}
Even-Dar, E., Mansour, Y., and Bartlett, P.
\newblock Learning rates for q-learning.
\newblock \emph{Journal of machine learning Research}, 5\penalty0 (1), 2003.

\bibitem[Focardi \& Fabozzi(2003)Focardi and Fabozzi]{focardi2003fat}
Focardi, S.~M. and Fabozzi, F.~J.
\newblock Fat tails, scaling, and stable laws: a critical look at modeling
  extremal events in financial phenomena.
\newblock \emph{The Journal of Risk Finance}, 2003.

\bibitem[Garg et~al.(2021)Garg, Zhanson, Parisotto, Prasad, Kolter,
  Balakrishnan, Lipton, Salakhutdinov, and Ravikumar]{garg2021proximal}
Garg, S., Zhanson, J., Parisotto, E., Prasad, A., Kolter, J.~Z., Balakrishnan,
  S., Lipton, Z.~C., Salakhutdinov, R., and Ravikumar, P.
\newblock On proximal policy optimization's heavy-tailed gradients.
\newblock \emph{arXiv preprint arXiv:2102.10264}, 2021.

\bibitem[Ghadimi \& Lan(2013)Ghadimi and Lan]{ghadimi2013stochastic}
Ghadimi, S. and Lan, G.
\newblock Stochastic first-and zeroth-order methods for nonconvex stochastic
  programming.
\newblock \emph{SIOPT}, 23\penalty0 (4):\penalty0 2341--2368, 2013.

\bibitem[Ghadimi et~al.(2016)Ghadimi, Lan, and Zhang]{ghadimi2016mini}
Ghadimi, S., Lan, G., and Zhang, H.
\newblock Mini-batch stochastic approximation methods for nonconvex stochastic
  composite optimization.
\newblock \emph{Mathematical Programming}, 155\penalty0 (1-2):\penalty0
  267--305, 2016.

\bibitem[Hutchinson(1981)]{hutchinson1981fractals}
Hutchinson, J.~E.
\newblock Fractals and self similarity.
\newblock \emph{Indiana University Mathematics Journal}, 30\penalty0
  (5):\penalty0 713--747, 1981.

\bibitem[Kakade(2002)]{kakade2002natural}
Kakade, S.~M.
\newblock A natural policy gradient.
\newblock In \emph{NeurIPS}, pp.\  1531--1538, 2002.

\bibitem[Khodadadian et~al.(2021)Khodadadian, Jhunjhunwala, Varma, and
  Maguluri]{khodadadian2021linear}
Khodadadian, S., Jhunjhunwala, P.~R., Varma, S.~M., and Maguluri, S.~T.
\newblock On the linear convergence of natural policy gradient algorithm.
\newblock \emph{arXiv preprint arXiv:2105.01424}, 2021.

\bibitem[Konda \& Borkar(1999)Konda and Borkar]{konda1999actor}
Konda, V.~R. and Borkar, V.~S.
\newblock Actor-critic--type learning algorithms for {M}arkov decision
  processes.
\newblock \emph{SICON}, 38\penalty0 (1):\penalty0 94--123, 1999.

\bibitem[Konda \& Tsitsiklis(2000)Konda and Tsitsiklis]{konda2000actor}
Konda, V.~R. and Tsitsiklis, J.~N.
\newblock Actor-critic algorithms.
\newblock In \emph{NeurIPS}, pp.\  1008--1014, 2000.

\bibitem[Kosorok \& Moodie(2015)Kosorok and Moodie]{kosorok2015adaptive}
Kosorok, M.~R. and Moodie, E.~E.
\newblock \emph{Adaptive treatment strategies in practice: planning trials and
  analyzing data for personalized medicine}.
\newblock SIAM, 2015.

\bibitem[Kushner \& Yin(2003)Kushner and Yin]{kushner2003stochastic}
Kushner, H.~J. and Yin, G.~G.
\newblock \emph{Stochastic approximation and recursive algorithms and
  applications}.
\newblock Springer, New York, NY, 2003.

\bibitem[Lan(2021)]{lan2021policy}
Lan, G.
\newblock Policy mirror descent for reinforcement learning: Linear convergence,
  new sampling complexity, and generalized problem classes.
\newblock \emph{arXiv preprint arXiv:2102.00135}, 2021.

\bibitem[Lillicrap et~al.(2016)Lillicrap, Hunt, Pritzel, Heess, Erez, Tassa,
  Silver, and Wierstra]{lillicrap2016continuous}
Lillicrap, T.~P., Hunt, J.~J., Pritzel, A., Heess, N., Erez, T., Tassa, Y.,
  Silver, D., and Wierstra, D.
\newblock Continuous control with deep reinforcement learning.
\newblock In \emph{International Conference on Learning Representations}, 2016.

\bibitem[Matheron et~al.(2019)Matheron, Perrin, and
  Sigaud]{matheron2019problem}
Matheron, G., Perrin, N., and Sigaud, O.
\newblock The problem with ddpg: understanding failures in deterministic
  environments with sparse rewards.
\newblock \emph{arXiv preprint arXiv:1911.11679}, 2019.

\bibitem[Nemirovski et~al.(2009)Nemirovski, Juditsky, Lan, and
  Shapiro]{nemirovski2009robust}
Nemirovski, A., Juditsky, A., Lan, G., and Shapiro, A.
\newblock Robust stochastic approximation approach to stochastic programming.
\newblock \emph{SIAM Journal on optimization}, 19\penalty0 (4):\penalty0
  1574--1609, 2009.

\bibitem[Nguyen et~al.(2019)Nguyen, {\c{S}}im{\c{s}}ekli,
  G{\"u}rb{\"u}zbalaban, and Richard]{nguyen2019first}
Nguyen, T.~H., {\c{S}}im{\c{s}}ekli, U., G{\"u}rb{\"u}zbalaban, M., and
  Richard, G.
\newblock First exit time analysis of stochastic gradient descent under
  heavy-tailed gradient noise.
\newblock \emph{arXiv preprint arXiv:1906.09069}, 2019.

\bibitem[Papini et~al.(2018)Papini, Binaghi, Canonaco, Pirotta, and
  Restelli]{papini2018stochastic}
Papini, M., Binaghi, D., Canonaco, G., Pirotta, M., and Restelli, M.
\newblock Stochastic variance-reduced policy gradient.
\newblock In \emph{ICML}, pp.\  4026--4035, 2018.

\bibitem[Papini et~al.(2020)Papini, Battistello, and
  Restelli]{papini2020balancing}
Papini, M., Battistello, A., and Restelli, M.
\newblock Balancing learning speed and stability in policy gradient via
  adaptive exploration.
\newblock In \emph{International Conference on Artificial Intelligence and
  Statistics}, pp.\  1188--1199. PMLR, 2020.

\bibitem[Puterman(2014)]{puterman2014markov}
Puterman, M.~L.
\newblock \emph{Markov \relax{D}ecision \relax{P}rocesses: \relax{D}iscrete
  stochastic dynamic programming}.
\newblock John Wiley \& Sons, 2014.

\bibitem[Rockafellar(1976)]{rockafellar1976monotone}
Rockafellar, R.~T.
\newblock Monotone operators and the proximal point algorithm.
\newblock \emph{SIAM journal on control and optimization}, 14\penalty0
  (5):\penalty0 877--898, 1976.

\bibitem[Schulman et~al.(2015)Schulman, Moritz, Levine, Jordan, and
  Abbeel]{schulman2015high}
Schulman, J., Moritz, P., Levine, S., Jordan, M., and Abbeel, P.
\newblock High-dimensional continuous control using generalized advantage
  estimation.
\newblock \emph{arXiv preprint arXiv:1506.02438}, 2015.

\bibitem[Schulman et~al.(2017)Schulman, Wolski, Dhariwal, Radford, and
  Klimov]{schulman2017proximal}
Schulman, J., Wolski, F., Dhariwal, P., Radford, A., and Klimov, O.
\newblock Proximal policy optimization algorithms.
\newblock \emph{arXiv preprint arXiv:1707.06347}, 2017.

\bibitem[Simsekli et~al.(2020{\natexlab{a}})Simsekli, Sener, Deligiannidis, and
  Erdogdu]{NEURIPS2020_37693cfc}
Simsekli, U., Sener, O., Deligiannidis, G., and Erdogdu, M.~A.
\newblock Hausdorff dimension, heavy tails, and generalization in neural
  networks.
\newblock In Larochelle, H., Ranzato, M., Hadsell, R., Balcan, M.~F., and Lin,
  H. (eds.), \emph{Advances in Neural Information Processing Systems},
  volume~33, pp.\  5138--5151. Curran Associates, Inc., 2020{\natexlab{a}}.

\bibitem[Simsekli et~al.(2020{\natexlab{b}})Simsekli, Zhu, Teh, and
  Gurbuzbalaban]{simsekli2020fractional}
Simsekli, U., Zhu, L., Teh, Y.~W., and Gurbuzbalaban, M.
\newblock Fractional underdamped langevin dynamics: Retargeting sgd with
  momentum under heavy-tailed gradient noise.
\newblock In \emph{International Conference on Machine Learning}, pp.\
  8970--8980. PMLR, 2020{\natexlab{b}}.

\bibitem[Sutton et~al.(2000)Sutton, McAllester, Singh, and
  Mansour]{sutton2000policy}
Sutton, R.~S., McAllester, D.~A., Singh, S.~P., and Mansour, Y.
\newblock Policy gradient methods for reinforcement learning with function
  approximation.
\newblock In \emph{NeurIPS}, pp.\  1057--1063, 2000.

\bibitem[Sutton et~al.(2017)Sutton, Barto, et~al.]{sutton2017reinforcement}
Sutton, R.~S., Barto, A.~G., et~al.
\newblock \emph{Reinforcement learning: {A}n introduction}.
\newblock 2 edition, 2017.

\bibitem[Tomar et~al.(2020)Tomar, Shani, Efroni, and
  Ghavamzadeh]{tomar2020mirror}
Tomar, M., Shani, L., Efroni, Y., and Ghavamzadeh, M.
\newblock Mirror descent policy optimization.
\newblock \emph{arXiv preprint arXiv:2005.09814}, 2020.

\bibitem[Watkins \& Dayan(1992)Watkins and Dayan]{watkins1992q}
Watkins, C.~J. and Dayan, P.
\newblock Q-learning.
\newblock \emph{Machine learning}, 8\penalty0 (3-4):\penalty0 279--292, 1992.

\bibitem[Williams(1992)]{williams1992simple}
Williams, R.~J.
\newblock Simple statistical gradient-following algorithms for connectionist
  reinforcement learning.
\newblock \emph{Machine Learning}, 8\penalty0 (3-4):\penalty0 229--256, 1992.

\bibitem[Xu et~al.(2019)Xu, Gao, and Gu]{xu2019sample}
Xu, P., Gao, F., and Gu, Q.
\newblock Sample efficient policy gradient methods with recursive variance
  reduction.
\newblock \emph{arXiv preprint arXiv:1909.08610}, 2019.

\bibitem[Xu et~al.(2017)Xu, Liu, and Peng]{xu2017stochastic}
Xu, T., Liu, Q., and Peng, J.
\newblock Stochastic variance reduction for policy gradient estimation.
\newblock \emph{arXiv preprint arXiv:1710.06034}, 2017.

\bibitem[Yang et~al.(2019)Yang, Zheng, Zhang, Zhang, Zheng, Wen, and
  Pan]{yang2019policy}
Yang, L., Zheng, G., Zhang, H., Zhang, Y., Zheng, Q., Wen, J., and Pan, G.
\newblock Policy optimization with stochastic mirror descent.
\newblock \emph{arXiv preprint arXiv:1906.10462}, 2019.

\bibitem[Yuan et~al.(2020)Yuan, Lian, Liu, and Zhou]{yuan2020stochastic}
Yuan, H., Lian, X., Liu, J., and Zhou, Y.
\newblock Stochastic recursive momentum for policy gradient methods.
\newblock \emph{arXiv preprint arXiv:2003.04302}, 2020.

\bibitem[Zhang et~al.(2020{\natexlab{a}})Zhang, Kim, O'Donoghue, and
  Boyd]{zhang2020sample}
Zhang, J., Kim, J., O'Donoghue, B., and Boyd, S.
\newblock Sample efficient reinforcement learning with reinforce.
\newblock \emph{arXiv preprint arXiv:2010.11364}, 2020{\natexlab{a}}.

\bibitem[Zhang et~al.(2020{\natexlab{b}})Zhang, Koppel, Zhu, and
  Basar]{zhang2020global}
Zhang, K., Koppel, A., Zhu, H., and Basar, T.
\newblock Global convergence of policy gradient methods to (almost) locally
  optimal policies.
\newblock \emph{SIAM Journal on Control and Optimization}, 58\penalty0
  (6):\penalty0 3586--3612, 2020{\natexlab{b}}.

\bibitem[Zou et~al.(2019)Zou, Xia, Ding, Song, Liu, and
  Yin]{zou2019reinforcement}
Zou, L., Xia, L., Ding, Z., Song, J., Liu, W., and Yin, D.
\newblock Reinforcement learning to optimize long-term user engagement in
  recommender systems.
\newblock In \emph{Proceedings of the 25th ACM SIGKDD International Conference
  on Knowledge Discovery \& Data Mining}, pp.\  2810--2818, 2019.

\end{thebibliography}
\bibliographystyle{icml2022}
\newpage


\onecolumn
\appendix
\normalsize
\section*{\centering Appendix}

\section{Proof of Lemma \ref{cauchy_bounded}}\label{bounded}
Since we consider Cauchy distribution, we note that the norm of the score function is given by
\begin{align}\label{second_gradient}
	\|\nabla\log \pi_{\bm{\theta}} (s,a)\|  = \frac{2\|(\|a-\varphi(s)^\top\bm{\theta}_{1}\|/{\sigma})}{1+ {(\|a-\varphi(s)^\top\bm{\theta}_{1}\|/{\sigma})^2 }} \left(  \frac{\|\varphi(s)\|}{\sigma} \right) ,
\end{align}
which is of the form $\frac{2x}{1+x^2}$ and it holds that $\frac{2x}{1+x^2}\leq 1$ for all $x$. Hence we conclude that $$\|\nabla\log \pi_{\bm{\theta}} (s,a)\|\leq \left(  \frac{\|\varphi(s)\|}{\sigma} \right) \leq \frac{D}{\sigma}:=B.$$
Hence proved. 
\section{Proof of Lemma \ref{lemtrack}}\label{proof_lemma}
\begin{proof}
	\normalfont
	The proof follows along the lines of \citep[Lemma 2]{cutkosky2019momentum}. Let us denote $\bar\delta_k=\nabla F(\bbtheta_k)$, expand the expression $\hbg_k- \delta_k$,  and introducing $(1-\beta)\bar\delta_{k-1}$, we obtain
	\begin{align}
		\hbg_{k} - \bar{\delta}_k &=  \left(1-\beta\right)(\hat \bbg_{k-1}-{\widetilde{\nabla} F(\bbtheta_{k-1},{\xi_k(\bbtheta_{k})})})+ \nabla F(\bbtheta_k,{\xi_k(\bbtheta_{k})})- \bar{\delta}_k
		\nonumber
		\\
		&=  \left(1-\beta\right)(\hat \bbg_{k-1}-\bar{\delta}_{k-1}+\bar{\delta}_{k-1}-{\widetilde{\nabla} F(\bbtheta_{k-1},{\xi_k(\bbtheta_{k})})}+ \nabla F(\bbtheta_k,{\xi_k(\bbtheta_{k})})- \bar{\delta}_k
		\nonumber
		\\
		&= (1-\beta)(\hbg_{k-1} - \bar{\delta}_{k-1})  - \big((1-\beta)[{\widetilde{\nabla} F(\bbtheta_{k-1},{\xi_k(\bbtheta_{k})})}-\bar{\delta}_{k-1}] - \nabla F(\bbtheta_{k},{\xi_k(\bbtheta_{k})}) + \bar{\delta}_k\big).
	\end{align}
	{Note that in the above expression, ${\xi_k(\bbtheta_{k-1})}$ denotes the trajectory collected at iteration $k$ using the policy parameter ${\bbtheta_{k-1}}$ and ${\xi_k(\bbtheta_{k})}$ denotes the trajectory collected at iteration $k$ using ${\bbtheta_{k}}$ as the policy parameter. Hence, this would imply that 	$\E{\bar{\delta}_{k-1} - {\widetilde{\nabla} F(\bbtheta_{k-1},{\xi_k(\bbtheta_{k})})}} = \E{\nabla F(\bbtheta_{k},\xi_k(\bbtheta_{k})) - \bar{\delta}_k} = 0$.} 
%
 Therefore, we have that
	\begin{align}
		\E{\norm{\hbg_{k}  - \bar{\delta}_k}^2} &= (1-\beta)^2\norm{\hbg_{k-1}  - \bar{\delta}_{k-1}}^2 \nonumber\\
		&+ \E{\norm{(1-\beta)({\widetilde{\nabla} F(\bbtheta_{k-1},{\xi_k(\bbtheta_{k})})}-\bar{\delta}_{k-1}) + \bar{\delta}_k - \nabla F(\bbtheta_{k},{\xi_k(\bbtheta_{k})}) }^2}\label{track2},
	\end{align}
	where the cross term vanishes since $\hbg_{k-1} - \bar{\delta}_{k-1}$ is independent of $\xi_k$ and the second summand is zero mean. Let us define $\e_k:=	\E{\norm{\hbg_{k}  - \bar{\delta}_k}^2}$, we could write \eqref{track2} as 
	%
\begin{align}
	\e_k &= (1-\beta)^2\e_{k-1} + \E{\norm{(1-\beta)({\widetilde{\nabla} F(\bbtheta_{k-1},{\xi_k(\bbtheta_{k})})}-\bar{\delta}_{k-1}) + \bar{\delta}_k - \nabla F(\bbtheta_{k},{\xi_k(\bbtheta_{k})}) }^2}\label{track22}.
\end{align}
The second term on the right hand side of \eqref{track2} can again be expanded as
	\begin{align}
		&\E{\norm{(1-\beta)({\widetilde{\nabla} F(\bbtheta_{k-1},{\xi_k(\bbtheta_{k})})}-\bar{\delta}_{k-1}) + \bar{\delta}_k - \nabla F(\bbtheta_{k},{\xi_k(\bbtheta_{k})}) }^2} \nonumber
		\\
		&\qquad= \E{\norm{(1-\beta)(\bar{\delta}_{k-1} - {\widetilde{\nabla} F(\bbtheta_{k-1},{\xi_k(\bbtheta_{k})})} + \nabla F(\bbtheta_{k},{\xi_k(\bbtheta_{k})}) - \bar{\delta}_k) + \beta(\nabla F(\bbtheta_{k},{\xi_k(\bbtheta_{k})}) - \bar{\delta}_k)}^2} 
		\\
		&\qquad \leq 2(1-\beta)^2\E{{\norm{\nabla F(\bbtheta_{k},{\xi_k(\bbtheta_{k})}) - {\widetilde{\nabla} F(\bbtheta_{k-1},{\xi_k(\bbtheta_{k})})}}^2}} + 2\beta^2\E{\norm{\nabla F(\bbtheta_{k},{\xi_k(\bbtheta_{k})})- \bar{\delta}_k}^2} \label{track3},
	\end{align}
	where we have used the inequality $\E{\norm{{X}-\E{{X}} + {Y}}^2} \leq 2\E{\norm{{X}}^2} + 2\E{\norm{{Y}}^2}$ for any random variables ${X}$ and ${Y}$ with bounded variances. 
	%
 Substitute Assumption \ref{assum:second1} regarding the gradient estimation error for the second term to obtain:
 	\begin{align}
		\e_{k} &\leq (1-\beta)^2\e_{k-1} + {2(1-\beta)^2\E{{\norm{\nabla F(\bbtheta_{k},{\xi_k(\bbtheta_{k})}) - {\widetilde{\nabla} F(\bbtheta_{k-1},{\xi_k(\bbtheta_{k})})}}^2}}}+ 2\beta^2m_0 {+2m_1\beta^2\|\nabla F(\bbtheta_{k})\|}^2.\label{track70}
	\end{align}
Next, we focus on the second term on the right hand side of \eqref{track7} as follows.
\begin{align}
& \E{{ 		\norm{\nabla F(\bbtheta_{k},{\xi_k(\bbtheta_{k})}) - {\widetilde{\nabla} F(\bbtheta_{k-1},{\xi_k(\bbtheta_{k})})}}^2}} 
\nonumber
\\
&\hspace{1cm}=\E{{ 		\norm{\nabla F(\bbtheta_{k},{\xi_k(\bbtheta_{k})}) - \nabla F(\bbtheta_{k-1},{\xi_k(\bbtheta_{k})})+\nabla F(\bbtheta_{k-1},{\xi_k(\bbtheta_{k})})- {\widetilde{\nabla} F(\bbtheta_{k-1},{\xi_k(\bbtheta_{k})})}}^2}}
\nonumber
\\
&\hspace{1cm}\leq 2\E{{ 		\norm{\nabla F(\bbtheta_{k},{\xi_k(\bbtheta_{k})}) - \nabla F(\bbtheta_{k-1},{\xi_k(\bbtheta_{k})})}^2}}+2\E{\norm{{\nabla F(\bbtheta_{k-1},{\xi_k(\bbtheta_{k})})- {\widetilde{\nabla} F(\bbtheta_{k-1},{\xi_k(\bbtheta_{k})})}}}^2}
\nonumber
\\
&\hspace{1cm}\leq 2L^2\E{{ 		\norm{\bbtheta_{k}-\bbtheta_{k-1}}^2}}+2\E{\norm{{\nabla F(\bbtheta_{k-1},{\xi_k(\bbtheta_{k-1})})- {\widetilde{\nabla} F(\bbtheta_{k-1},{\xi_k(\bbtheta_{k})})}}}^2}. \label{track10}
\end{align}
The second term in the right hand side of \eqref{track10} is unique to the policy gradient settings. The importance weights based definition of gradient would help us to bound this term. Let us expand the second term on the right hand side of \eqref{track10} as follows
\begin{align}
&	2\E{\norm{{\nabla F(\bbtheta_{k-1},{\xi_k(\bbtheta_{k-1})})- {\widetilde{\nabla} F(\bbtheta_{k-1},{\xi_k(\bbtheta_{k})})}}}^2}
	\\
	&\hspace{1cm} = 2\E{\Big\|{{\sum_{t=0}^{T_k}\gamma^{t/2}r_t\cdot\bigg(\sum_{\tau=0}^{t}\nabla\log\pi_{\bm{\theta}_{k-1}}(a_{\tau}\given s_{\tau})\bigg)- \sum_{t=0}^{T_k}\gamma^{t/2}r_t\cdot w_t\left({\tau~|~\bbtheta_{k-1},\bbtheta_{k}}\right)\big(\sum_{\tau=0}^{t}\nabla\log\pi_{\bm{\theta}_{k-1}}(a_{\tau}\given s_{\tau})\big)}}\Big\|^2} \nonumber.
\end{align}
After collecting the like terms together, we obtain
\begin{align}
	&	2\E{\norm{{\nabla F(\bbtheta_{k-1},{\xi_k(\bbtheta_{k-1})})- {\widetilde{\nabla} F(\bbtheta_{k-1},{\xi_k(\bbtheta_{k})})}}}^2}
	\\
	&\hspace{1cm} = 2\E{\Big\|{{\sum_{t=0}^{T_k}\gamma^{t/2}r_t\cdot\left(1-w_t\left({\tau~|~\bbtheta_{k-1},\bbtheta_{k}}\right)\right)\cdot\bigg(\sum_{\tau=0}^{t}\nabla\log\pi_{\bm{\theta}_{k-1}}(a_{\tau}\given s_{\tau})\bigg)}}\Big\|^2} \nonumber
		\\
	&\hspace{1cm} = 2\E{\sum_{t=0}^{T_k}\mathbb{E}\left[\Big\|{{\gamma^{t/2}r_t\cdot\left(1-w_t\left({\tau~|~\bbtheta_{k-1},\bbtheta_{k}}\right)\right)\cdot\bigg(\sum_{\tau=0}^{t}\nabla\log\pi_{\bm{\theta}_{k-1}}(a_{\tau}\given s_{\tau})\bigg)}}\Big\|^2\right]}, 
\end{align}
The second equality holds due to the fact that $\mathbb{E}\left[\nabla\log\pi_{\bm{\theta}}(a\given s)\right]=0$.  Further, utilizing the upper bounds on reward and absolute bound on the score function norm, we can write
\begin{align}
	&	2\E{\norm{{\nabla F(\bbtheta_{k-1},{\xi_k(\bbtheta_{k-1})})- {\widetilde{\nabla} F(\bbtheta_{k-1},{\xi_k(\bbtheta_{k})})}}}^2}
	\\
	&\hspace{1cm} \leq  2U_R^2\E{\sum_{t=0}^{T_k}\mathbb{E}\left[\gamma^{t}\cdot\Big\|{{\left(1-w_t(\tau~|~\bbtheta_{k-1},\bbtheta_{k})\right)\Big\|^2\cdot t\sum_{\tau=0}^{t}\Big\|\nabla\log\pi_{\bm{\theta}_{k-1}}(a_{\tau}\given s_{\tau})}}\Big\|^2\right]}
		\\
	&\hspace{1cm} \leq  2U_R^2\E{\sum_{t=0}^{T_k}\mathbb{E}\left[\gamma^{t}\cdot\Big\|{{\left(1-w_t(\tau~|~\bbtheta_{k-1},\bbtheta_{k})\right)\Big\|^2\cdot t\sum_{\tau=0}^{t}\Big\|\nabla\log\pi_{\bm{\theta}_{k-1}}(a_{\tau}\given s_{\tau})}}\Big\|^2\right]}		\\
	&\hspace{1cm} \leq  2U_R^2\E{\sum_{t=0}^{T_k}\mathbb{E}\left[\gamma^{t}\cdot\Big\|{{\left(1-w_t(\tau~|~\bbtheta_{k-1},\bbtheta_{k})\right)\Big\|^2\cdot t^2B^2}}\right]}
	\\
	&\hspace{1cm} \leq  2U_R^2B^2\E{\sum_{t=0}^{T_k}\gamma^{t}t^2\mathbb{E}\left[\cdot\Big\|{{\left(1-w_t(\tau~|~\bbtheta_{k-1},\bbtheta_{k})\right)\Big\|^2\cdot }}\right]}
		\\
	&\hspace{1cm} =  2U_R^2B^2\E{\sum_{t=0}^{T_k}\gamma^{t}t^2\text{Var}\left(w_t(\tau~|~\bbtheta_{k-1},\bbtheta_{k})\right)},
\end{align}
which holds because $\mathbb{E}[w_t(\tau~|~\bbtheta_{k-1},\bbtheta_{k})]=1$ \citep[Lemma C.1]{xu2019sample}. Next, utilizing Assumption \ref{importance_variance} and \citep[Lemma B.1]{xu2019sample}, we can write
\begin{align}
		2\E{\norm{{\nabla F(\bbtheta_{k-1},{\xi_k(\bbtheta_{k-1})})- {\widetilde{\nabla} F(\bbtheta_{k-1},{\xi_k(\bbtheta_{k})})}}}^2}
	&\hspace{0cm} \leq  2U_R^2B^2C_w\E{\sum_{t=0}^{T_k}\gamma^{t}t^2\mathbb{E}\left[\|\bbtheta_{k}-\bbtheta_{k-1}\|^2\right]}
	\\
	&\hspace{0cm} \leq  2U_R^2B^2C_w\E{T_k^2\mathbb{E}\left[\|\bbtheta_{k}-\bbtheta_{k-1}\|^2\right]\sum_{t=0}^{T_k}\gamma^{t}}
	\\
	&\hspace{0cm} \leq  \frac{2U_R^2B^2C_w}{1-\gamma}\E{\|\bbtheta_{k}-\bbtheta_{k-1}\|^2 \mathbb{E}\left[T_k^2\right]}.
\end{align}
Since $T_k\sim\text{Geom}(p)$ with $p=1-\gamma^{1/2}$, it holds that $\EE[T_k^2]=(\EE[T_k])^2 + var(T_k)$, which further implies that $\EE[T_k^2]=\frac{1+\sqrt{\gamma}}{(1-\sqrt{\gamma})^2}$. Hence, ww can write
\begin{align}
	&	2\E{\norm{{\nabla F(\bbtheta_{k-1},{\xi_k(\bbtheta_{k-1})})- {\widetilde{\nabla} F(\bbtheta_{k-1},{\xi_k(\bbtheta_{k})})}}}^2} \leq  \frac{2U_R^2B^2C_w(1+\sqrt{\gamma})}{(1-\gamma)((1-\sqrt{\gamma})^2)}\E{\|\bbtheta_{k}-\bbtheta_{k-1}\|^2}. \label{last}
\end{align}
 Utilizing the upper bound in \eqref{last} into \eqref{track10}, we obtain
\begin{align}
	& \E{{ 		\norm{\nabla F(\bbtheta_{k},{\xi_k(\bbtheta_{k})}) - {\widetilde{\nabla} F(\bbtheta_{k-1},{\xi_k(\bbtheta_{k})})}}^2}} \leq L_1\E{{ 		\norm{\bbtheta_{k}-\bbtheta_{k-1}}^2}}, \label{track11}
\end{align}
where we define $L_1:=\left(2L^2+\frac{2U_R^2B^2C_w(1+\sqrt{\gamma})}{(1-\gamma)((1-\sqrt{\gamma})^2)}\right)$. Hence, using \eqref{track11} into \eqref{track70}, we obtain
\begin{align}
	\e_{k} &\leq (1-\beta)^2\e_{k-1} + {2(1-\beta)^2\eta^2L_1\E{{ 		\norm{\bbtheta_{k}-\bbtheta_{k-1}}^2}}}+ 2\beta^2m_0 {+2m_1\beta^2\|\nabla F(\bbtheta_{k})\|}^2\nonumber
.\label{track7}
\end{align}
From the definition of generalized Bregman gradient, we can write
\begin{align}
	\e_{k} &\leq (1-\beta)^2\e_{k-1} + {2(1-\beta)^2L_1\E{{\|\mathcal{G}_{\eta,\hbg_k}^\psi(\bbtheta_k)\|^2}}}+ 2\beta^2m_0 {+2m_1\beta^2\|\nabla F(\bbtheta_{k})\|}^2	\\
	&\leq (1-\beta)^2\e_{k-1} + {2\eta^2L_1\E{{ 		\norm{\bbtheta_{k}-\bbtheta_{k-1}}^2}}}+ 2\beta^2m_0 {+2m_1\beta^2\|\nabla F(\bbtheta_{k})\|}^2,\label{track77}
\end{align}
	where \eqref{track77} follows from substituting in the difference of policy parameters by the generalized gradient in \eqref{modified}, and the second expression uses the fact that $\beta\leq 1$.
\end{proof}

\section{Proof of Theorem \ref{main_khorem}}\label{proof_theorem}

\begin{proof}
	\normalfont
%
From the smoothness of $F(\cdot)$ [cf. Assumption \ref{ass:smooth}], it holds that
\begin{align}
	F(\bbtheta_{k+1})\leq F(\bbtheta_{k})+\langle\nabla F(\bbtheta_k), \bbtheta_{k+1}-\bbtheta_{k}\rangle +\frac{L}{2}\|\bbtheta_{k+1}-\bbtheta_{k}\|^2. \label{first}
\end{align}
From the update in \eqref{modified} (note that the analysis is for descent update), it holds that
\begin{align}
	\bbtheta_{k+1} = &\bbtheta_{k} - \eta  \mathcal{G}_{\eta,\hat\bbg_k}^\psi(\bm{\theta}_{k}).\label{udpate2}
\end{align}
Let us utilize \eqref{udpate2} into \eqref{first}, we get
\begin{align}
	F(\bbtheta_{k+1})\leq F(\bbtheta_{k})-\eta \langle\nabla F(\bbtheta_k), \mathcal{G}_{\eta,\hat\bbg_k}^\psi(\bm{\theta}_{k})\rangle +\frac{L\eta^2}{2}\|\mathcal{G}_{\eta,\hat\bbg_k}^\psi(\bm{\theta}_{k})\|^2. \label{first2}
\end{align}
Next, we add subtract the gradient $\hat \bbg_k$ as follows
\begin{align}
	F(\bbtheta_{k+1})\leq  &F(\bbtheta_{k})-\eta\langle\nabla F(\bbtheta_k)-\hat \bbg_k+\hat \bbg_k, \mathcal{G}_{\eta,\hat\bbg_k}^\psi(\bbtheta_k)\rangle +\frac{L\eta_k^2}{2}\|\mathcal{G}_{\eta,\hat\bbg_k}^\psi(\bbtheta_k)\|^2 \label{first3}
	\\
	=&F(\bbtheta_{k})-\eta\langle\hat \bbg_k, \mathcal{G}_{\eta,\hbg_k}^\psi(\bbtheta_k)\rangle +\frac{L\eta_k^2}{2}\|\mathcal{G}_{\eta,\hbg_k}^\psi(\bbtheta_k)\|^2+\eta\langle \bbw_k, \mathcal{G}_{\eta,\hbg_k}^\psi(\bbtheta_k)\rangle\label{first4}
\end{align}
where we utilized the definition $\bbw_k:=\hbg_k-\nabla F(\bbtheta_k)$ is the stochastic error in the gradient in the second line. Next, using the lower bound from \eqref{prop1}, we can substitute the second term by the square-norm of the Bregman gradient as 
%
%
\begin{align}
	F(\bbtheta_{k+1})\leq  & F(\bbtheta_{k})-\eta\zeta\|\mathcal{G}_{\eta,\hbg_k}^\psi(\bbtheta_k)\|^2 +\frac{L\eta_k^2}{2}\|\mathcal{G}_{\eta,\hbg_k}^\psi(\bbtheta_k)\|^2+\eta\langle \bbw_k, \mathcal{G}_{\eta,\hbg_k}^\psi(\bbtheta_k)\rangle\label{first6}.
\end{align}
%

Next, for the last inner product term on the right hand side of \eqref{first6}, we use $ab\leq \beta a^2+\frac{1}{\beta}b^2$ for any $\beta>0$ with $a= \bbw_k$ and $b=\eta \mathcal{G}_{\eta,\hbg_k}^\psi(\bbtheta_k)$ to rewrite the preceding expression as
\begin{align}
	F(\bbtheta_{k+1})\leq  & F(\bbtheta_{k})-\eta\zeta\|\mathcal{G}_{\eta,\hbg_k}^\psi(\bbtheta_k)\|^2 +\beta\|\bbw_k\|^2+\frac{\eta^2}{\beta}\|\mathcal{G}_{\eta,\hbg_k}^\psi(\bbtheta_k)\|^2+\frac{L\eta^2}{2}\|\mathcal{G}_{\eta,\hbg_k}^\psi(\bbtheta_k)\|^2\label{first62}.
\end{align}
Next, after grouping like terms, we get 
\begin{align}
	F(\bbtheta_{k+1})\leq  & F(\bbtheta_{k})-\eta\zeta\left(1-\frac{\eta}{\beta \zeta}\right)\|\mathcal{G}_{\eta,\hbg_k}^\psi(\bbtheta_k)\|^2 +\frac{L\eta^2}{2}\|\mathcal{G}_{\eta,\hbg_k}^\psi(\bbtheta_k)\|^2+\beta\|\bbw_k\|^2\label{first63}.
\end{align}
By selecting $\beta\geq \frac{2\eta}{\zeta}$, it holds that,
\begin{align}
	F(\bbtheta_{k+1})\leq  & F(\bbtheta_{k})-\frac{\eta\zeta}{2}\|\mathcal{G}_{\eta,\hbg_k}^\psi(\bbtheta_k)\|^2 +\frac{L\eta^2}{2}\|\mathcal{G}_{\eta,\hbg_k}^\psi(\bbtheta_k)\|^2+\beta\|\bbw_k\|^2\label{first64}.
\end{align}
Taking the expectation on the both sides, we get
\begin{align}
	\mathbb{E}	[F(\bbtheta_{k+1})]\leq  & \mathbb{E}[F(\bbtheta_{k})]-\frac{\eta\zeta}{2}\mathbb{E}\left[\|\mathcal{G}_{\eta,\hbg_k}^\psi(\bbtheta_k)\|^2\right] +\frac{L\eta^2}{2}\mathbb{E}\left[\|\mathcal{G}_{\eta,\hbg_k}^\psi(\bbtheta_k)\|^2\right]+\beta\mathbb{E}\left[\|\bbw_k\|^2\right]\label{first65}.
\end{align}
%

%
%
Before proceeding next, let us provide an upper bound on the term $\mathbb{E}\left[\|\bbw_k\|^2\right]$ in the form of Lemma \ref{lemtrack}.

From the result in \eqref{first65} and the statement of Lemma \ref{lemtrack}, we write
\begin{align}
	\e_{k+1} \leq&  (1-\beta)^2\e_{k} + 2\eta^2 {L_{t_{\text{mix}}}}{\mathbb{E}\left[\|\mathcal{G}_{\eta,\hbg_k}^\psi(\bbtheta_k)\|^2\right]} + 2\beta^2m_0 {+2m_1\beta^2\|\nabla F(\bbtheta_{k})\|}^2\label{track0}
	\\
	\mathbb{E}	[F(\bbtheta_{k+1})]\leq  & \mathbb{E}[F(\bbtheta_{k})]-\frac{\eta\zeta}{2}\mathbb{E}\left[\|\mathcal{G}_{\eta,\hbg_k}^\psi(\bbtheta_k)\|^2\right] +\frac{L\eta^2}{2}\mathbb{E}\left[\|\mathcal{G}_{\eta,\hbg_k}^\psi(\bbtheta_k)\|^2\right]+\beta\e_{k}\label{first67}.
\end{align}
Adding \eqref{track0} and \eqref{first67} then yields
\begin{align}
	\mathbb{E}	[F(\bbtheta_{k+1})]+\e_{k+1} \leq  & \mathbb{E}[F(\bbtheta_{k})]+\e_{k}-\e_{k}-\frac{\eta\zeta}{2}\mathbb{E}\left[\|\mathcal{G}_{\eta,\hbg_k}^\psi(\bbtheta_k)\|^2\right] +{\frac{L\eta^2}{2}\mathbb{E}\left[\|\mathcal{G}_{\eta,\hbg_k}^\psi(\bbtheta_k)\|^2\right]}+\beta\e_{k}\\
	& +(1-\beta)^2\e_{k} + 2\eta^2 {L_{t_{\text{mix}}}}{\mathbb{E}\left[\|\mathcal{G}_{\eta,\hbg_k}^\psi(\bbtheta_k)\|^2\right]} + 2\beta^2m_0{+2m_1\beta^2\mathbb{E}\left[\|\nabla F(\bbtheta_{k})\|^2\right]}\label{first68}.
\end{align}
Define Lyapunov function $\Phi_{k}:=\EE\left[F(\bbtheta_{k})\right]+	\e_{k}$. With this definition, after rearranging the preceding expression, we then obtain
\begin{align}
	\Phi_{k+1}- 	\Phi_{k}+ \frac{\eta\zeta}{2}\mathbb{E}\left[\|\mathcal{G}_{\eta,\hbg_k}^\psi(\bbtheta_k)\|^2\right] \leq  &  A_k\e_{k}+\frac{5\eta^2L'{\mathbb{E}\left[\|\mathcal{G}_{\eta,\hbg_k}^\psi(\bbtheta_k)\|^2\right]}}{2} + 2\beta^2m_0{+2m_1\beta^2\mathbb{E}\left[\|\nabla F(\bbtheta_{k})\|^2\right]}, \label{first23}
\end{align}
$L'=\max\{L,{L_{t_{\text{mix}}}}\}$ and $A_k=(1-\beta)^2+\beta-1$. Next, our goal is to establish that $\Phi_{k}$ is decreasing. To do so, we study the coefficient $A_k$ in front of $\varepsilon_k$ in more detail
\begin{align}
	A_k=&1+\beta^2-2\beta+\beta-1\\
	=&-\beta(1-\beta).
\end{align}
Note that since $\beta\leq 1$, we have $A_k\leq 0$ for all $t$, which allows us to drop the first term on the right-hand side of \eqref{first23} and hence write
\begin{align}
	\left(\frac{\eta\zeta}{2}-\frac{5\eta^2}{2L'}\right)\mathbb{E}\left[\|\mathcal{G}_{\eta,\hbg_k}^\psi(\bbtheta_k)\|^2\right] \leq  & 	\Phi_{k}-\Phi_{k+1}+ 2\beta^2m_0{+2m_1\beta^2\mathbb{E}\left[\|\nabla F(\bbtheta_{k})\|^2\right]}, \label{first242}
\end{align}
Next we use the error-bound condition on the second moment of the gradient ${\mathbb{E}\left[\|\nabla F(\bbtheta_{k})\|^2\right]}$ in Assumption \ref{assum:third1} as follows
\begin{align}
	{\mathbb{E}\left[\|\nabla F(\bbtheta_{k})\|^2\right]}=& {\mathbb{E}\left[\|\nabla F(\bbtheta_{k})-\mathcal{G}_{\eta,\hbg_k}^\psi(\bbtheta_k)+\mathcal{G}_{\eta,\hbg_k}^\psi(\bbtheta_k)\|^2\right]}
	\\
	\leq & {2\mathbb{E}\left[\|\nabla F(\bbtheta_{k})-\mathcal{G}_{\eta,\hbg_k}^\psi(\bbtheta_k)\|^2\right]}+{2\mathbb{E}\left[\|\mathcal{G}_{\eta,\hbg_k}^\psi(\bbtheta_k)\|^2\right]}
	\\
	\leq & {2m_2  +(2+m_3)\|\mathcal{G}_{\eta,\hbg_k}^\psi(\bbtheta_k)\|}^2.\label{upper_bound_norm}
\end{align}
Utilizing the upper bound of \eqref{upper_bound_norm} into the right hand side of \eqref{first242}, we get
\begin{align}
	\left(\frac{\eta\zeta}{2}-\frac{5\eta^2}{2L'}\right)\mathbb{E}\left[\|\mathcal{G}_{\eta,\hbg_k}^\psi(\bbtheta_k)\|^2\right] \leq  & 	\Phi_{k}-\Phi_{k+1}+ 2\beta^2m_0+{4m_1m_2\beta^2  +\tilde m_3\beta^2\mathbb{E}\left[\|\mathcal{G}_{\eta,\hbg_k}^\psi(\bbtheta_k)\|^2\right]}, \label{first2422220}
\end{align}
where $\tilde m_3=2(2+m_3)$.
By selecting $\eta\leq \frac{\zeta L'}{10}$ [{first condition on $\eta$}], we can lower bound the left hand side of \eqref{first2422220}, as
\begin{align}
	\frac{\eta\zeta}{4}\mathbb{E}\left[\|\mathcal{G}_{\eta,\hbg_k}^\psi(\bbtheta_k)\|^2\right] \leq  & 	\Phi_{k}-\Phi_{k+1}+ 2\beta^2m_0+{4m_1m_2\beta^2  +\tilde m_3\beta^2\mathbb{E}\left[\|\mathcal{G}_{\eta,\hbg_k}^\psi(\bbtheta_k)\|^2\right]}. \label{first2422220_new2}
\end{align}
After rearranging the terms, we get
\begin{align}
	\left(\frac{\eta\zeta}{4}-\tilde m_3 \beta^2\right)\mathbb{E}\left[\|\mathcal{G}_{\eta,\hbg_k}^\psi(\bbtheta_k)\|^2\right] \leq  & 	\Phi_{k}-\Phi_{k+1}+ 2\beta^2m_0+4m_1m_2\beta^2. \label{first2422220_new3}
\end{align}
Let use select $\beta=C_1\eta$ where $C_1$ is such that $C_1>\frac{2}{\zeta}$ which would satisfy our requirement for $\beta$. This implies that
\begin{align}
	\frac{\eta\zeta}{8}\left(2-\frac{8\tilde m_3 C_1^2\eta}{\zeta} \right)\mathbb{E}\left[\|\mathcal{G}_{\eta,\hbg_k}^\psi(\bbtheta_k)\|^2\right] \leq  & 	\Phi_{k}-\Phi_{k+1}+ 2\beta^2m_0+4m_1m_2\beta^2.  \label{first2422220_new4}
\end{align}
%
 Again if we make sure that $\eta\leq\frac{\zeta}{8\tilde m_3 C_1^2}$ [{second condition for $\eta$}], we could lower bound the above expression as
\begin{align}
	\frac{\eta\zeta}{8}\mathbb{E}\left[\|\mathcal{G}_{\eta,\hbg_k}^\psi(\bbtheta_k)\|^2\right] \leq  & 	\Phi_{k}-\Phi_{k+1}+ 2\beta^2m_0+4m_1m_2\beta^2. \label{first2422220_new5}
\end{align}
 Taking summation and lower bounding the left hand side with the minimum, we will obtain the similar rate as

 \begin{align}
 	\min_{1\leq k\leq K}\mathbb{E}\left[\|\mathcal{G}_{\eta,\hbg_k}^\psi(\bbtheta_k)\|^2\right]\leq & 	\mathcal{O}\left(\frac{1}{\eta K}+\eta\right),\label{first242222_new}
 \end{align}
where we used $\beta=C_1\eta$ and absorb the constants in  the $\mathcal{O}$ notation. 
 { From the above expression, by selecting $\eta =\epsilon$, we will obtain $K=\mathcal{O}(\epsilon^{-2})$. The explicit values of step size by collecting the conditions for $\eta$ and $\beta$ together could be written as 
 \begin{align}
 	\beta= C_1\eta\ \ \text{and}\ \  	\eta=  \min\Big\{\frac{\zeta L'}{10},\frac{\zeta}{8\tilde{m}_3 C_1^2}\Big\}.
\end{align} 
Similarly, if we choose, $\eta$$=$$\epsilon^2$, then from the Jensen's inequality, it would holds that $ 	\mathbb{E}\left[\|\mathcal{G}_{\eta,\hbg_k}^\psi(\bbtheta_k)\|\right]\leq \sqrt{\mathbb{E}\left[\|\mathcal{G}_{\eta,\hbg_k}^\psi(\bbtheta_k)\|^2\right]} \leq \epsilon $ with $K=\mathcal{O}\left(\frac{1}{\epsilon^4}\right)$ .
%
}
\end{proof}


\section{Verification of  Assumptions \ref{assum:first1}-\ref{ass:smooth} for the RL Setting} \label{assumptions_Satisfied}

\begin{enumerate}
	\item  Utilizing the boundedness of the score function as mentioned in Lemma \ref{cauchy_bounded}, combined with the proof of \citep[Lemma 1]{bedi2021sample}, it holds that Assumption \ref{assum:first1} is satisfied. 
	
	\item Assumption \ref{assum:second1} is regarding the variance of the stochastic gradient estimate which we assume to be bounded by some constant $m_0$ with $m_1=0$. This assumption is standard in the literature (see \cite{xu2017stochastic,xu2019sample,papini2018stochastic,zhang2020global}).
	
	\item To prove Assumption \ref{assum:third1}, we start by nothing that $\nabla F(\bbtheta)=\mathbb{E}[\nabla F(\bbtheta,\xi)]$ because stochastic gradient estimator is unbiased. Now we take norm on both sides, we get $\|\nabla F(\bbtheta)\| = \|\mathbb{E}[\nabla  F(\bbtheta,\xi)]\|$. Since norm is convex, we know that $\|\nabla F(\bbtheta)\| \leq  \mathbb{E} \|\nabla  F(\bbtheta,\xi)\|$. For the RL settings, from \eqref{eq:policy_gradient_iteration}, we know that
	\begin{align}
		{\nabla}  F(\bm{\theta},\xi)= \sum_{t=0}^{T}\gamma^{t/2}r_t\cdot\bigg(\sum_{\tau=0}^{t}\nabla\log\pi_{\bm{\theta}}(a_{\tau}\given s_{\tau})\bigg).
	\end{align}
Taking norm on both sides, applying triangle inequality, and utilizing the bound on reward and norm of the score function, we note that 

\begin{align}
	\|{\nabla}  F(\bm{\theta},\xi)\|\leq  \frac{U_R B}{1-\gamma} T.
\end{align}
Since $T\sim\text{Geom}(1-\gamma^{1/2})$, after taking expectation, we can write that  
\begin{align}
	\mathbb{E}\left[\|{\nabla}  F(\bm{\theta},\xi)\|^2\right]\leq  \frac{U_R B(1+\sqrt{\gamma})}{(1-\gamma)(1-\sqrt{\gamma})^2}.
\end{align}
From the state of Assumption \ref{assum:third1}, we note that term $\mathbb{E}\left[\|\nabla F(\bbtheta_k)\!-\! \mathcal{G}_{\eta,\hbg_k}^\psi(\bbtheta_k)\|^2\right]$ and use upper bound it as 
\begin{align}
	\mathbb{E}\left[\|\nabla F(\bbtheta_k)\!-\! \mathcal{G}_{\eta,\hbg_k}^\psi(\bbtheta_k)\|^2\right]\leq 	2\mathbb{E}\left[\|\nabla F(\bbtheta_k)\|^2\right]+ 2\mathbb{E}\left[\|\mathcal{G}_{\eta,\hbg_k}^\psi(\bbtheta_k)\|^2\right]. 
\end{align}			
Hence, Assumption \ref{assum:third1} holds with $m_2=\frac{2U_R B(1+\sqrt{\gamma})}{(1-\gamma)(1-\sqrt{\gamma})^2}$ and $m_3=2$. 

\item Since with the Cauchy policy parametrization, the score function is Lipschitz with parameter $L^{\pi}=\frac{2D^2}{\sigma^2}+\frac{7D}{\sigma^2}+1$. Using the Lipschitz property of the score function, along with the boundedness of the score function $\| \nabla \log \pi_{\bm{\theta}} (a|s)\|\leq B$, it holds that the objective $F$ is smooth with constant $L$ (proof is provided in \citep[Lemma 3.2]{zhang2020global}) and $L$ is given by
\begin{align}
L=	\frac{U_R L^{\pi}}{(1-\gamma)^2} + \frac{(1+\gamma)U_R B^2}{(1-\gamma)^3}.
\end{align}
\end{enumerate}

\end{document}